\documentclass[11pt, a4]{article}
\usepackage[hyperref]{emnlp2020}
\usepackage{times}
\usepackage{latexsym}

\usepackage{microtype}

\aclfinalcopy % Uncomment this line for the final submission
\usepackage{bm}      % microtypography
\usepackage{amsmath}
\usepackage{amsfonts}

\usepackage{tabularx}
\usepackage{subcaption}
\usepackage{todonotes}
\usepackage{mathtools}
\usepackage{comment}
\usepackage{breqn}

  \usepackage{multirow}
\usepackage{float}
\usepackage[toc,page]{appendix}
\usepackage{subfiles} % Best loaded last in the preamble

\newcommand{\cut}[1]{}
\newcommand{\xhdr}[1]{\noindent{\bfseries #1}.}

\title{TeMP: Temporal Message Passing for Temporal\\ Knowledge Graph Completion}

\author{%
  Jiapeng Wu \qquad Meng Cao \qquad Jackie Chi Kit Cheung \qquad William L. Hamilton\\
  School of Computer Science, McGill University, Montreal, QC, Canada \\
    MILA, Montreal, QC, Canada \\ \\
  \tt \{jiapeng.wu@mail, meng.cao@mail, jcheung@cs, wlh@cs\}.mcgill.ca\\
}

\begin{document}

\maketitle

\begin{abstract}
Inferring missing facts in temporal knowledge graphs (TKGs) is a fundamental and challenging task.
Previous works have approached this problem by augmenting methods for static knowledge graphs to leverage time-dependent representations.
However, these methods do not explicitly leverage multi-hop structural information and temporal facts from recent time steps to enhance their predictions. Additionally, prior work does not explicitly address the temporal {\em sparsity} and {\em variability} of entity distributions in TKGs.
We propose the \textbf{Te}mporal \textbf{M}essage \textbf{P}assing (TeMP) framework to address these challenges by combining graph neural networks, temporal dynamics models, data imputation and frequency-based gating techniques.
Experiments\footnote{Code and data are published at \url{https://github.com/JiapengWu/TeMP}} on standard TKG tasks show that our approach provides substantial gains compared to the previous state of the art, achieving a 10.7\% average relative improvement in Hits@10 across three standard benchmarks.
Our analysis also reveals important sources of variability both within and across TKG datasets, and we introduce several simple but strong baselines that outperform the prior state of the art in certain settings.
\end{abstract}

\section{Introduction}

The ability to infer missing facts in temporal knowledge graphs is essential for applications such as event prediction \citep{leblay2018deriving, de2018combining}, question answering \citep{jia2018tequila}, social network analysis \citep{zhou2018dynamic, trivedi2019dyrep} and recommendation systems \citep{kumar2018learning}. 

\begin{figure}
\centering
  \includegraphics[width=1
  \linewidth]{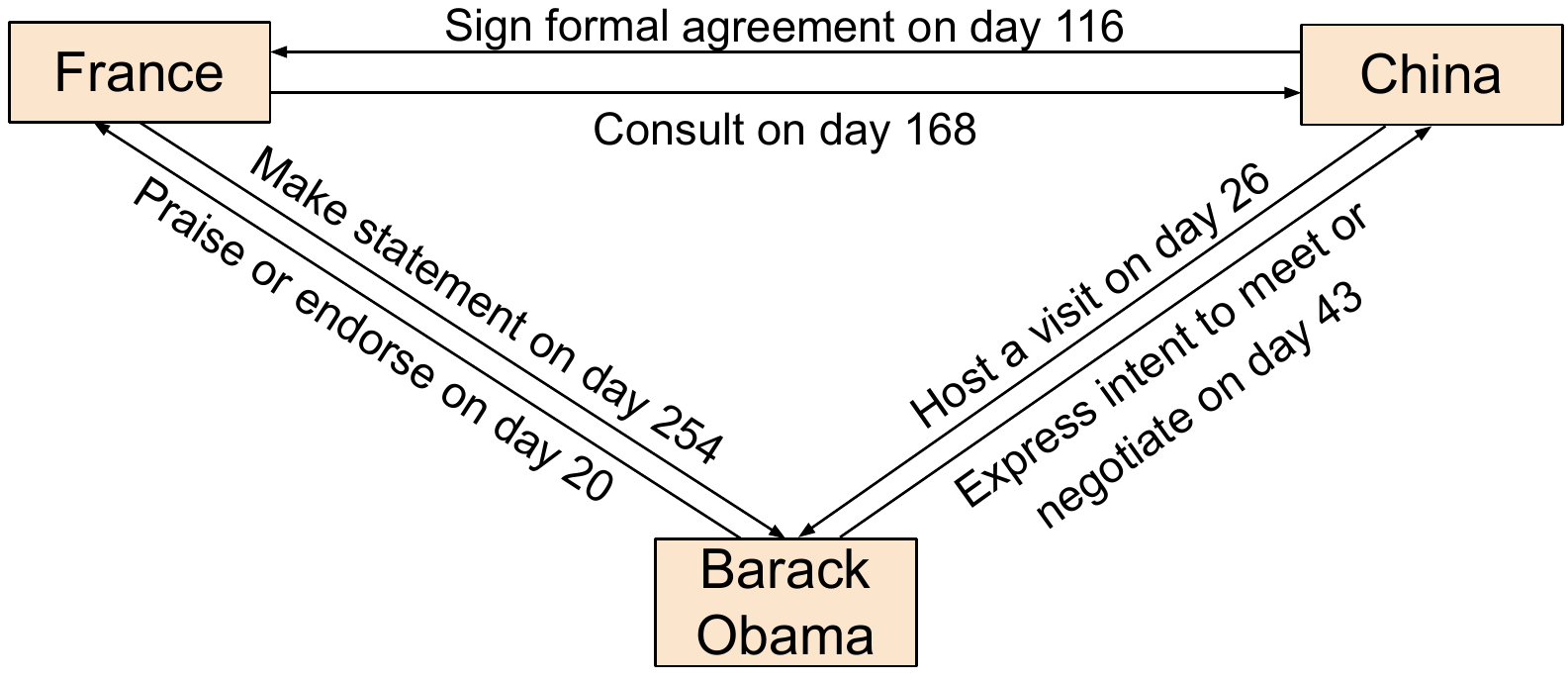} 
\caption{
A sample temporal knowledge subgraph involving France, China and Barack Obama.
}
\label{example_tkg}
\end{figure}

Whereas static knowledge graphs (KGs) represent facts as triples (e.g., \emph{(Obama, visit, China)}), temporal knowledge graphs (TKGs) additionally associate each triple with a timestamp (e.g., \emph{(Obama, visit, China, 2014)}). Figure \ref{example_tkg} shows a subgraph of such TKG. 
Usually, TKGs are assumed to consist of discrete timestamps \cite{jiang2016towards}, meaning that they can be represented as a sequence of static KG snapshots, and the task of inferring missing facts across these snapshots is referred to as temporal knowledge graph completetion (TKGC). 

Recent works on TKGC have largely focused on developing time-dependent scoring functions, which score the likelihood of missing facts and build closely upon popular representation learning methods for static KGs \citep{dasgupta2018hyte, jiang2016towards, goel2019diachronic, xu2019temporal, lacroix2020tensor}.
However, while powerful, these existing methods do not properly account for multi-hop structural information in TKGs, and they lack the ability to explicitly leverage temporal facts in nearby KG snapshots to answer queries. Knowing facts like \textit{(Obama, make agreement with, China, 2013)} or \textit{(Obama, visit, China, 2012)} is useful for answering the query \textit{(Obama, visit, ?, 2014)}.

Moreover---and perhaps more importantly---there are also serious challenges regarding {\em temporal variability} and {\em temporal sparsity}, which previous works fail to address.
In real-world TKGs, models have access to {\em variable} amounts of reference temporal information in near KG snapshots when answering different queries (Figure \ref{num_existing_hist_entities} and Figure \ref{num_existing_hist_entities_appendix} in the Appendix).
% and the sparsity of this historical data varies drastically across different entities and time steps (Figure \ref{num_existing_hist_entities}). 
For example, in a political event dataset, there are likely to be more quadruples with subject-relation pair \textit{(Obama, visit)} than \textit{(Trump, visit)} from 2008 to 2013.\footnote{Obama was the president of US during the period.} Hence the model could access more reference information to answer \textit{where Obama visited in 2014}.

The \emph{temporal sparsity} problem reveals that only a small fraction of entities are \emph{active}\footnote{An entity is \emph{active} at a time step if it has at least one neighboring entity in the same KG snapshot.} 
at each time step (Figure~\ref{num_avg_active_hist_nodes} in the Appendix).
Previous methods usually assign the same embedding for \emph{inactive} entities at different time steps, which is not fully representative of the time-sensitive features.
%More detailed analysis of these issues are presented in Appendix \ref{section dataset statistics}. 
%Will: I moved this sentence to the caption

% We carry out more detailed analysis regarding temporal variability and sparsity problem in appendix \todo{refer to appendix}.

% Will: Some of this content should be moved to the figure caption. 
%Figure \ref{num_existing_hist_entities} shows the temporal distribution of \emph{active} entities on ICEWS14 \cite{garcia2018learning} dataset.  While the total number of entities in ICEWS14 is 7281, only 2\% -- 4\% of these entities are active at each time step in ICEWS14. Among the active entities in some snapshots, only about half of them have occurred at least once in the last 15 time steps.
%More detailed analysis of these issues are presented in Appendix \ref{section dataset statistics}. 

% \item Diminishing effect of historical entity representation with respect to the increase temporal difference to the current time step. For example, the entity representation of \emph{Obama} and \emph{China} in the year 2013 is more useful for inferring the given fact than their representations in the year 2000. 

% We alleviate the variability problem by developing attention-style gating of static and temporal entity representation.
% In this work, we make the following contributions: 

\xhdr{Present work}
To address these issues, we introduce the \textbf{Te}mporal \textbf{M}essage \textbf{P}assing (TeMP) framework, which combines neural message passing and temporal dynamic models. 
We then propose frequency-based gating and data imputation techniques to counter the temporal sparsity and variability issues. 
% While these techniques improve the model performances on ICEWS datasets, GDELT does not benefit from these. 

% The variabilities across TKGs cause the different effect of these techniques 

% We explore the effect of these techniques on different datasets and found a source of difference in these problems that caused different effect of these techniques.
We achieve state-of-the-art performance on standard TKGC benchmarks.
In particular, on the standard ICEWS14, ICEWS05-15, and GDELT datasets, TeMP is able to provide an 7.3\% average relative improvement in Hits@10 compared to the next-best model. 
% Experimental study suggests the influence of dataset statistics on the effect of each individual technique.
Fine-grained error analysis on these three datasets demonstrates the unique contributions made by each of the different components of TeMP. 
Our analysis also highlights important sources of variability, in particular  variations in temporal sparsity both within and across TKG datasets, and how effects of different components are affected by such variability.
% Our analysis also highlights important variability both within and across TKGs, and demonstrates how the effects of different components of TeMP are affected by such variability. 

%I would mention the code in the experiments section
%Our code is released at ...\todo{upload zipped code files to a web url}

% We then develop several mechanisms to tackle the aforementioned heterogeneities in TKG. We propose data imputation for sparsity problem and alleviate the variability problem by applying gating of static and temporal entity representation. 
% \begin{itemize}
    % \item We achieved state-of-the-arts performance with the combination of message passing network and temporal dependency model. 
    % \item We propose exponential decay mechanism to better model diminishing effect of events and tackle the sparsity of entity occurrence on the time dimension.
    % \item We alleviated the imbalance of the amount of reference information by developing attention-style gating of static and temporal entity representation.
    % \item 
% \end{itemize}

% Existing TKG embedding and static RGCN behave relatively invariant of such quantities while there exists clear positive correlation among temporal pattern frequencies and model performance for temporal models. To combine the advantages of static RGCN and temporal RGCN over different temporal pattern frequencies, we proposed an gating method based on the query type and entity positions. 

%\todo[inline]{include code code repo}

\begin{figure}
\centering
  \includegraphics[width=0.7
  \linewidth]{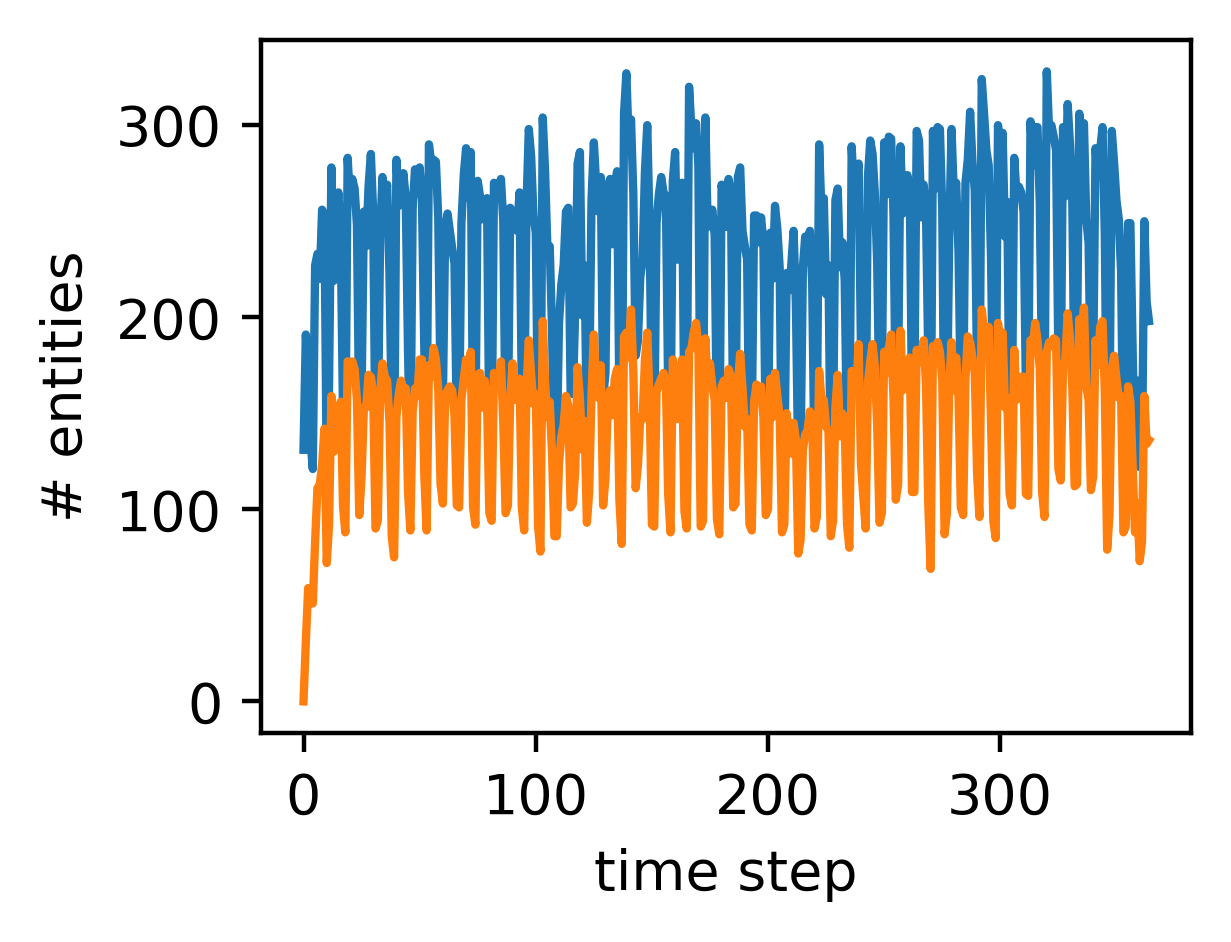} 
\caption{Dataset statistics of the ICEWS14 dataset. The blue (top) curve shows the number of active entities at each time step, while the orange (bottom) curve represents the number of active entities at each time step that are also active at least once in the past 15 time steps. 
% The sharp increase at the beginning is the result of the increasing number of available preceding time steps. 
While the total number of entities is 7,128, only 2\% -- 4\% of these entities are active at each time step.
(See Appendix \ref{section dataset statistics} for further examples and discussion). 
}
\label{num_existing_hist_entities}
\end{figure}

% Our approach in this work differs to theirs in that  1) we provide a much simplified model that leverages historical node information at different layers of convolution 2) we apply exponential decaying function that down-weights the historic features.

\section{Related Work}

\paragraph{Static KG representation learning}

Much research exists on representation learning methods for static KGs, in which entities and relations are represented as low-dimensional embeddings \citep{nickel2011three, yang2014embedding,trouillon2016complex, nickel2016holographic}. 
%These works focus on a series of knowledge graph completion tasks. 
%In these works, entities and relations are represented by low dimensional vectors or matrices, and various decoding functions are proposed to score the triples. 
Generally, these methods involve a {\em decoding} method, which scores candidate facts based on entity and relation embeddings, and the models are optimized so that valid triples receive higher scores than random {\em negative examples}.
While these methods typically rely on shallow encoders to generate the embeddings---i.e., single embedding-lookup layers \cite{hamilton2017representation}---message passing (or graph neural network; GNN) approaches have also been proposed \cite{schlichtkrull2018modeling, vashishth2019composition, busbridge2019relational} to leverage multi-hop information around entities. 
%have also been proposed, which encodes multi-relational data using the differentiable message passing mechanism. 
% The representation of an entity is computed after applying relation-specific transformations on its neighbors for multiple iterations. Our work leverages the representation capacity of such mechanism to model structural features within each KG snapshot. 

\paragraph{Temporal KG representation learning}
Recent works endeavor to extend static KGC models to the temporal domain. Typically, such approaches employ embedding methods with a shallow encoder and design time-sensitive quadruple decoding functions \citep{dasgupta2018hyte, jiang2016towards, goel2019diachronic, xu2019temporal, lacroix2020tensor}. While time-specific information is considered by these methods, entity-level temporal patterns such as event periodicity are not explicitly captured. 
% The TKG embedding methods explicitly learn time-aware entity and relation representations through time-dependent projection or gating and incorporate them into the entity-relation space in KG.
% The effects of concurrent events in snapshots of at different time are is omitted in the reasoning process. This type of information is crucial and heavily exploited in the discrete time KG. 

Another line of work on temporal (knowledge) graph reasoning uses message passing networks to capture intra-graph neighborhood information, which is sometimes combined with temporal recurrence or attention mechanisms \citep{manessi2020dynamic, kumar2018learning, pareja2019evolvegcn, chen2018gc, jin2019recurrent, sankar2020dysat, hajiramezanali2019variational}.
%Typically, these methods employ recurrence and self-attention mechanism to model the evolution of nodes and graph. In recurrent models, the hidden states of RNN is used to represent aggregated graph information at each time step. 
Orthogonal to our work, \citet{trivedi2017know, trivedi2019dyrep, han2020graph} explore using temporal point processes. However, their focus is on continuous TKGC.
The prior works that most resemble our framework are Recurrent Event Networks (RE-NET) \citep{jin2019recurrent} and DySAT \cite{sankar2020dysat}. RE-NET uses multi-level RNNs to model entity interactions, while DySAT uses self-attention to learn latent node representations on dynamic graphs. 
However, both these works were proposed for the task of graph extrapolation (i.e., inferring the next time-step in a sequence), so they are not directly compatible with the TKGC setting. 
%The learned entity representations of both models are used for \emph{extrapolation} problem, i.e. prediction future interactions, which is not compatible with the TKGC setting.

%\subfile{sections/background}

\section{Proposed Approach}

We first define our key notation and provide an overview of our TeMP framework, before describing the individual components in detail in the following sections. 

\xhdr{Notation and task definition}
Our goal is to predict missing facts in a temporal knowledge graph (TKG) $\mathcal{G} = \{G^{(1)}, G^{(2)}, ..., G^{(T)}\}$, where $G^{(t)} = (E, R, D^{(t)})$.
Here, $E$ and $R$ stand for the union of sets of entities and relations across all time steps and are known in advance. $D^{(t)}$ denotes the set of all \emph{observed} triples $(s, r, o)$ at time $t$, with subjects $s \in E$ , objects $ o \in E$ and relations $r \in R$. 
Let $\overline{D}^{(t)}$ denote the set of \emph{true} triples at time $t$ such that $D^{(t)} \subseteq \overline{D}^{(t)}, \forall t$, the temporal knowledge graph completion (TKGC) problem is defined as ranking the subject and object entities given object queries $(s, r, ?, t)$ and subject queries $(?, r, o, t)$ where $(s, r, o) \in \overline{D}^{(t)}$ but $(s, r, o) \not\in D^{(t)}$, $t \in \{0, ... ,T\}$. \\

\xhdr{Overview of TeMP}

Following common practice, we structure our TeMP framework around the notion of an {\em encoder} and {\em decoder}.
The encoder maps each entity $e_i \in E$ to time-dependent low-dimensional embedding $\bm{z}_{i,t}$ at each time-step $t$, while the decoder uses these entities' embeddings to score the likelihood of a temporal fact.
%Note that unlike previous work, we leverage a {\em time-dependent encoder}, but our decoder model does not explicitly depend on time.

\begin{figure}
\centering
  \includegraphics[width=1
  \linewidth]{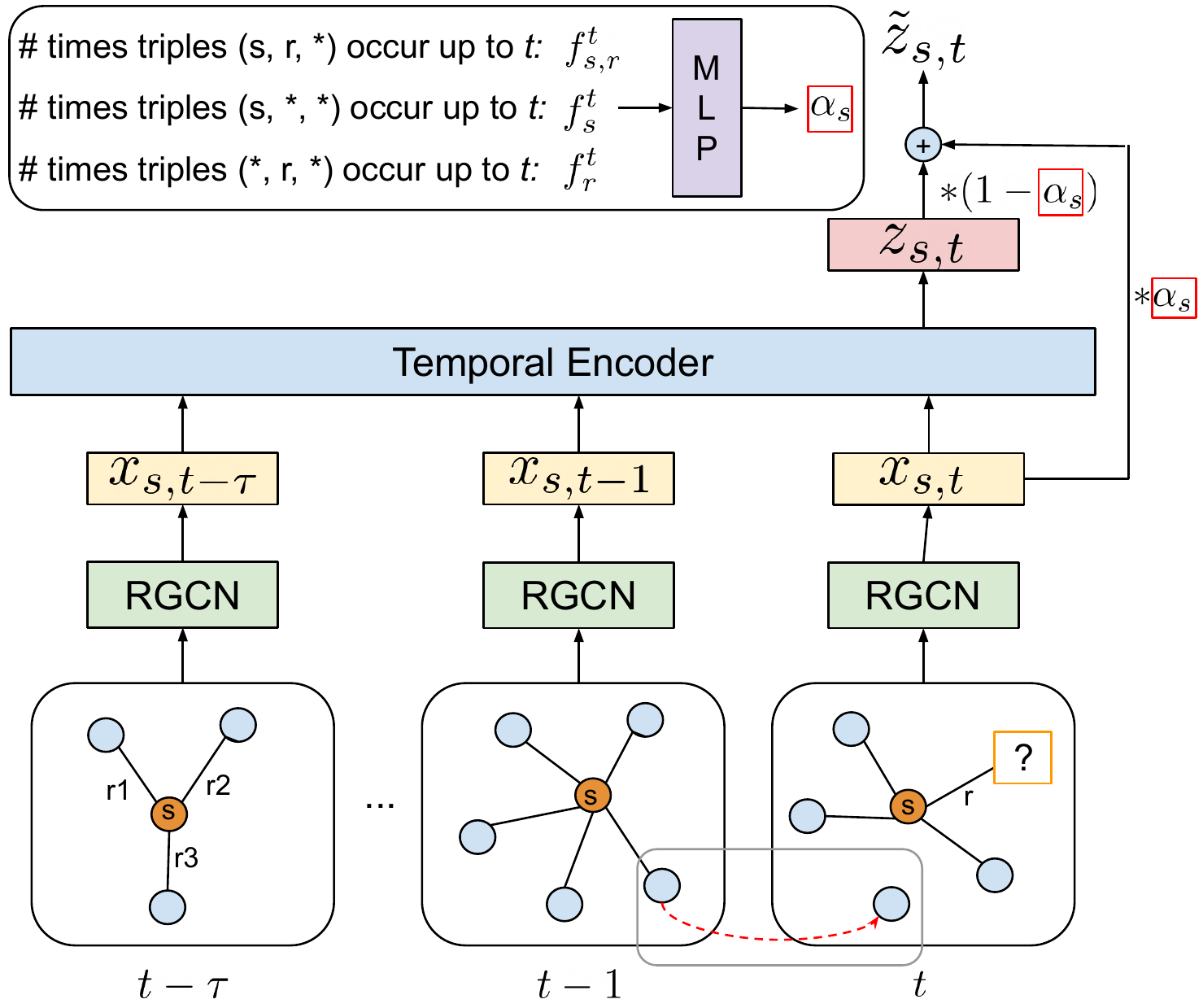} 
\caption{
Architecture of TeMP Framework. TeMP combines structural graph encoder and temporal encoder to induce entity representations. Given query $(s, r, ?, t)$ at time $t$, TeMP takes graphs from time step $t-\tau$ to $t$ as input to compute structural embedding $\bm{x}_{s, t}$ and temporal embedding $\bm{z}_{s, t}$ for the centering entity $s$. 
The final representation $\bm{\tilde{z}}_{s, t}$ is obtained by further applying frequency-based gating, as illustrated in the upper rectangle. The red dotted arrow at the bottom indicates the imputation process for an inactive entity at time step $t$.
}
\label{architecture_figure}
\end{figure}

Figure \ref{architecture_figure} depicts the architecture of our model. A key insight in TeMP is that we use an encoder that combines a  \emph{structural} entity representation and \emph{temporal} representations. The structural encoder (SE) based on a multi-relational message passing network produces entity representation $\bm{x}_{i,t} = \text{SE}(e_i, D^{(t)})$ while the temporal encoder ($\text{TE}$) integrates the output of SE at previous time steps to induce $\bm{z}_{i,t} = \text{TE}(\bm{x}_{i,t - \tau}, ..., \bm{x}_{i,t})$. Here $\tau$ stands for the number of temporal input KG snapshots to the model.

% At time step $t$, we use structural encoder to induce $\mb{x}_{i,t}$ and use temporal encoder to get $\mb{z}_{i,t}$. These two components are introduced in Sections \ref{sec:structural} and \ref{sec:temporal}, respectively. 
In addition, in Section \ref{sec:heterogeneity}, we propose a series of augmentations to TeMP that are designed to address the temporal sparsity and variability issues of real-world TKGs.
Finally, in Section \ref{sec:training}, we discuss how TeMP can leverage existing decoders from the static KG setting in order to train a model.

\subsection{Structural Encoder}\label{sec:structural}

The first key component of TeMP is the structural encoder, which generates entity embeddings based on the graph $G^{(t)}$ within each time-step.
We build our structural encoder by adapting existing techniques for message passing on static knowledge graphs \cite{schlichtkrull2018modeling}. 

% We focus on RGCN as encoder. However, our framework is not tied to any specific multi-relational message passing network. A naive extension of RGCN to temporal domain without introducing new parameters is to share the entity embedding and model weights across all time steps. 
% The message-passing process is defined as follows:
% Namely for the $i$-th entity occurring in the $t$-th snapshot, 

\begin{align*}
\bm{h}_{i,t}^{(0)} &= \bm{W}_0 \bm{u}_i, \forall t \in {0, ..., T},\\
\bm{h}_{i, t}^{(l+1)} &= \sigma\Big(\sum_{r \in R} \sum_{j \in N_{i}^r} \frac{1}{|N_{i}^r|} \bm{W}_r^{(l)}\bm{h}_{j, t}^{(l)} + \bm{W}_s^{(l)}\bm{h}_{i, t}^{(l)} \Big)
\end{align*}

Here, $\bm{u}_i$ denotes a one-hot embedding indicating entity $e_i$, $\bm{W}_0$ is an entity embedding matrix, and $\bm{W}_r^{(l)}$ and $\bm{W}_s^{(l)}$ are transformation matrices specific to each layer of the model. These matrices are shared across all discrete time stamps.  
We use $N_{i}^r$ to denote the set of neighboring entities of $e_i$ connected by relation $r$, whose size acts as a normalizing constant for averaging the neighborhood information.
After running $L$ layers of this message-passing approach on a snapshot $G^{(t)}$, we use $\bm{x}_{i,t} = \bm{h}_{i,t}^{(L)}$ to denote the resulting structural embedding of entity $e_i$, which summarizes its $L$-hop neighborhood within $G^{(t)}$.

While we focus on RGCN as the structural encoder, our framework is not tied to any specific multi-relational message passing network.  One can swap RGCN with any multi-relational graph encoder, e.g. CompGCN \citep{vashishth2019composition} and EdgeGAT \citep{busbridge2019relational}.
%The static entity representation is denoted as $\bm{x}_{i,t} = \bm{h}_{i,t}^{(L)}$, the output of $L$-th relation convolution.  

\subsection{Temporal Encoder}\label{sec:temporal}

The second key component of TeMP is the temporal encoder, which seeks to integrate information \emph{across} time in the entity representations.
We investigate two approaches to compute entity representation $\bm{z}_{i, t}$ leveraging temporal information: a recurrent architecture (inspired by \citet{jin2019recurrent}) and a self-attention approach (inspired by \citet{sankar2020dysat}). 

%Let $t$ denote current the time step.
%We use $\tau$ to denote size of the sliding  the number of KG snapshots within the time window surrounding $G^{(t)}$ to encode the current graph. An entity $e$ is \emph{active} at time $t$ if $\exists (s, r, o, t) \in D^{(t)}$, $s = e$ or $o = e$. We use $\bm{z}_{i,t}$ to denote the entity representation of the $i$-th entity at the $t$-th time step. 

\xhdr{Temporal recurrence model (TeMP-GRU)}\label{recurrence}
%Recurrent mechanism has been widely used for modelling sequential data. 
We propose to couple a traditional recurrence mechanism with weight decay, in order to account the diminishing effect of historical facts.
%Intuitively, the effect of historical entity representation decreases with as the temporal distance to the current time step increases. For example, the entity representation of \emph{Obama} and \emph{China} in the year 2013 is more useful for inferring the $(Obama, visits, ?, 2014)$ than their representations in the year of 2000.  
%In particular, we use an exponential decay function to downweight the historical representation and set the decay rate as a learnable parameter. 
Let $t^-$ denote the last time step at which entity $e_i$ was \emph{active} before $t$, the down-weighted entity representation $\hat{\bm{z}}_{i,t^{-}}$ is defined as follows:
\begin{align}
\bm{\hat{z}}_{i,t^{-}} &= \gamma_{i, t^-}^z \bm{z}_{i, t^{-}}\label{eq:expdecay}\\ 
\gamma_{i, t^-}^z &= \exp\{-\max(0, \lambda_z|t-t^-| + b_z)\},
\end{align}
where $\gamma^z$ denotes the decay rate with $\lambda_z$ and $b_z$ as learnable parameters. This design is inspired by \citet{che2018recurrent} and ensures that $\gamma^z$ is monotonically decreasing with respect to the temporal difference and ranges from 0 to 1. We ensure that $\bm{\hat{z}}_{i,t^-}$ is only nonzero if $t^- \in \{t-\tau,.., t-1\}$, otherwise it will be assigned a zero vector. 
Finally, we use a gated recurrent unit (GRU) to obtain the entity embedding $\bm{z}_{i,t}$ based on $\bm{\hat{z}}_{i,t^-}$ and the static representation $\bm{x}_{i, t}$:
\begin{equation}\label{RRGCN_3}
\bm{z}_{i,t} = \text{GRU}\big(\bm{x}_{i, t} , \bm{\hat{z}}_{i,t^-}\big),
\end{equation}
where GRU denotes the standard cell defined by \citet{cho2014learning}.

% In the experiment we also consider layer-wise temporal recurrence to incorporate historical information summarizing neighbors within different number of hops in the previous graphs to the corresponding current entity representation.

\xhdr{Temporal self attention model (TeMP-SA)}\label{sttention}
Another way to incorporate historical information is to selectively attend to the sequence of active temporal entity representations.
%\cite{sankar2020dysat} found that temporal self-attention layer captures temporal variations in uni-relation graph structure, we hypothesis that self-attention mechanism also benefit TKGC task by dynamically incorporating temporal entity representations. 
We use the following equations---inspired by the transformer architecture \cite{vaswani2017attention}---to perform attentive pooling over the entity embeddings $\bm{x}_{i,t'}$ at each time step $t' \in \{t-\tau,.., t\}$, in order to generate time-dependent embeddings $\bm{z}_{i,t}$:
\begin{align}
&\bm{q}_{ij} = \frac{(\bm{x}_{i,t} \bm{W}_q)(\bm{x}_{i,t-j} \bm{W}_k)^T}{\sqrt{d}}\label{equation:attention_1}\\
&\bm{e}_{ij} = \bm{q}_{ij} - \max(0, \lambda_z j + b_z) + \bm{M}_{ij}\\
&\bm{\beta}_{ij} = \frac{\exp(\bm{e}_{ij})}{\sum_{k=0}^\tau \exp(\bm{e}_{ik})}\\
&\bm{z}_{i,t} = \sum_{j=0}^{\tau} \bm{\beta}_{ij}(\bm{x}_{i,t-j}\bm{W}_v)\label{equation:attention_4},
%&\bm{z}_{i,t} = \text{Concat}(\bm{a}_{i,t}^1, \bm{a}_{i,t}^2,...,\bm{a}_{i,t}^{H})
\end{align}
where $\bm{W}_q, \bm{W}_k, \bm{W}_v\in \mathbb{R}^{d \times d} $ denote linear projection matrices, as in a transformer layer \citep{vaswani2017attention}, $\bm{\beta} \in \mathbb{R}^{|E| \times \tau}$ denotes the attention weight matrix obtained by multiplicative attention function and $\{\lambda_z,b_z\}$ denotes the learnable parameters of the down-weighting function.
The $\bm{M} \in \mathbb{R}^{|E| \times \tau}$ matrix is a mask defined as
\begin{equation}\label{SARGCN_3}
\bm{M}_{ij} = 
\begin{cases}
  0, & \text{if $e_i$ is active at time $t-j$,}\\
  -\infty,  & \text{otherwise.}
\end{cases}
\end{equation}

As $\bm{M}_{ij} \rightarrow {-\infty}$, the attention weights $\bm{\beta}_{ij} \rightarrow 0$, which ensures that only active temporal entity representations are assigned non-zero weights. 
Finally, note that the full self-attention model can be generalized to use multiple attention heads, as in \citet{vaswani2017attention}. 

%We employ multi-head attention mechanism to model the multi-faceted KG evolution:
%\begin{equation}\label{SARGCN_4}
%\bm{z}_{i,t} = \text{Concat}(\bm{a}_{i,t}^1, %\bm{a}_{i,t}^2,...,\bm{a}_{i,t}^{H})
%\end{equation}
%here $H$ denotes the number of attention heads. 

\xhdr{Incorporating future information}
Note that in the TKGC setting, we assume that the model has access to all the time steps during training.
In particular, we assume there is missing data within each time step but that all the (incomplete) snapshots information $D^{(t)}$ are available during training. 
Thus, in both the attention and recurrence-based approaches, it is worthwhile to integrate temporal information from both the past and future. 
We do so by employing a bi-directional GRU in the recurrent approach, and by attending over both past and future time steps in the attention-based approach.

\subsection{Tackling Temporal Heterogeneities}\label{sec:heterogeneity}
% The structural and temporal encoders allow us to generate entity embeddings that leverage both multi-hop structural information as well as temporal information from nearby snapshots. 
Although TeMP jointly models structural and temporal information, the encoder alone is insufficient to deal with the \emph{temporal heterogeneity} in real-world TKGs, namely \emph{sparsity} and \emph{variability} of entity occurrences.
We explore  data imputation and frequency-based gating techniques to address these temporal heterogeneities. Because the degrees of temporal heterogeneities vary drastically across datasets (Appendix \ref{section dataset statistics}), our proposed techniques are optional model variations that may improve model performance depending on the dataset characteristics.

\xhdr{Imputation of inactive entities}
% In addition to frequency-based gating, we also implement an approach to perform data imputation to address the temporal sparsity problem. 
% We notice that in some scenarios only a small subset of entities are active in each KG snapshots 
Recall that structural encoder only encodes neighboring entities within the same KG snapshot. For entity $e_i$ that is inactive at time step $t$, the static representation $\bm{x_{i, t}}$ is hence not informed by any structural neighbors, resulting in stale representations shared across multiple time steps. 
%This can be seen as a form of informative missingness in the KG domain. There exist several methods tackling such issue in time series modelling, we use an approach shown to be effective in \cite{che2018recurrent} to \textit{impute} the representation of non-active entities. 
% Inspired by \cite{che2018recurrent}, we propose an approach to {\em impute} the representation of inactive entities. 
We propose an imputation (IM) approach that integrates stale representations with temporal representations for inactive entities, i.e., $\bm{\hat{x}}_{i, t} = \text{IM}(\bm{x}_{i,t}, \bm{x}_{i,t-})$, where $\bm{\hat{x}}_{i, t}$ represents imputed structural representation.

Without loss of generality, we define the imputation for a uni-directional model and refer the bidirectional case to Appendix \ref{sec:extended imputation}. We defined IM to be the weighted sum function, with the similar exponential decay mechanism used in Equation~\eqref{eq:expdecay}:
\begin{equation}\label{equation:imputation_1}
\gamma_{i,t^-}^x = \exp\{-\max(0, \lambda_x|t-t^-| + b_x)\}.
\end{equation}
The imputed representation is defined as follows:
\begin{equation}\label{equation:imputation_2}
    \bm{\hat{x}}_{i, t} = \gamma_{i,t^-}^x \bm{x}_{i, t^-} + (1 - \gamma_{i,t^-}^x) \bm{x}_{i, t}.
\end{equation}
This model-agnostic approach is applicable by replacing $\bm{x}_{i, t}$ in the temporal models with $\bm{\hat{x}}_{i, t}$ .
% with $\bm{\hat{x}}_{i, t}$ before being used by temporal encoder. 

\xhdr{Frequency-based gating}
% In addition to frequency-based gating, we also implement an approach to perform data imputation to address the temporal sparsity problem. 
In addition to imputation, we also implement an approach to perform frequency-based gating (FG). The encoded representation of an entity is modulated depending on how many recent temporal facts it participates in. 
In particular, we propose to learn a gating term in order to fuse the embeddings $\bm{x}_{i,t}$ from output of the structural encoder (Section \ref{sec:structural}) with the temporal embeddings $\bm{z}_{i,t}$ (Section \ref{sec:temporal}) in a frequency-dependent way. We differentiate the weights by the query types (subject or object query) and entity position (whether $e_i$ is subject or object in the queried fact) in order to contextualize the entities into their role within a quadruple.

In what follows, we use the term {\em pattern} to denote a non-empty subset of the quadruple $(s, r, o, t)$ (not containing time $t$). 
% For example, patterns associate with $(Obama, visit, China, 2014)$ include \textit{Obama}, \textit{China}, \textit{(Obama, visit)}, and \textit{(visit, China)}.
The \textit{temporal frequency} of a pattern is defined as the number of facts with such pattern in the defined time window. Consider the quadruple \textit{(Obama, visit, China, 2014)}, the temporal frequency of the pattern \textit{(Obama, visit)} is the number of quadruples \textit{(Obama, visit, $*$, $t'$)} with $t'$ in the time window (e.g., from 2000 to 2014). 
% A pattern is a subset of the triple $(s, r, o)$, e.g. $s$, $o$, $\{s, r\}$, $\{s, r, o\}$. The temporal frequency of a pattern is defined as the number of triples with such pattern in the defined time window. 
% i.e. $\{t-\tau,.., t+\tau\}$ in bidirectional setting and $\{t-\tau,.., t-1\}$ in vanilla setting. 

We define the following \emph{temporal pattern frequencies} (TPFs) associated with the quadruple $(s, r, o, t)$: (1) subject frequency $f_s^t$, (2) object frequency $f_o^t$, (3) relation frequency $f_r^t$, (4) subject-relation frequency $f_{s,r}^t$, (5) relation-object frequency $f_{r,o}^t$.

Without loss of generality, we define our gating mechanism from the perspective of object queries $(s, r, ?, t)$, where the goal is to predict the missing object in a quadruple. 
The definition for subject queries is analogous and detailed in Appendix \ref{analogous gating}.

When answering the object query $(s, r, ?, t)$ the model has only the access to frequencies $F_s = [f_s^t, f_r^t, f_{s, r}^t]$.
Thus, we use the frequency vector $F_s$ to define a gating term over the embeddings in the query:
\begin{align}
\bm{\tilde{z}}_{s, t} & = \alpha_{os} \bm{x}_{s, t} + (1 - \alpha_{os}) \bm{z}_{s, t}\label{eq:gating-1}\\
\bm{\tilde{z}}_{o, t} & = \alpha_{oo} \bm{x}_{o, t} + (1 - \alpha_{oo}) \bm{z}_{o, t}\label{eq:gating-2},
\end{align}
where $\alpha_{os} = \text{MLP}_{os}(F_s), \alpha_{oo} = \text{MLP}_{oo}(F_s)$ are weights in the range $[0,1]$ learned via a two-layer dense neural network. Here the calculation for object embedding $\bm{\tilde{z}}_{o, t}$ covers all entities.

\subsection{Decoder and Training}\label{sec:training}
Let $\phi(.)$ denote the score for a tuple and let DEC denote any proper decoding function for static KGs, e.g., the TransE decoder \citep{bordes2013translating}.
The score for the quadruple is defined as follows:
\begin{equation}\label{decoder}
    \phi(s, r, o, t)= \text{DEC}(\bm{\tilde{z}}_{s, t}, \bm{z}_r, \bm{\tilde{z}}_{o, t}).
\end{equation}
Here, $\bm{\tilde{z}}_{s,t}$ and $\bm{\tilde{z}}_{o,t}$ are the subject and object embeddings (as defined in Sections \ref{sec:structural}-\ref{sec:heterogeneity}) while $\bm{z}_r$ is a learned embedding of the relation $r$.
To train a model using this score function, the model parameters are learned using gradient-based optimization in mini-batches. 
For each triple $\eta = (s, r, o) \in D^{(t)}$, we sample a negative set of entities $D^-_{\eta} = \{o' | (s, r, o') \not\in D^{(t)}\}$ and define the cross-entropy loss as follows:
\begin{equation*}\label{loss_obj}
L = - \sum_{t=1}^T\sum_{\eta \in D^{(t)}}\frac{\exp(\phi(s, r, o, t))}{\sum_{o' \in D^-_{\eta}} \exp(\phi(s, r, o', t))}.
\end{equation*}
Note that without loss of generality, we defined the above loss over object queries (as in Section \ref{sec:heterogeneity}), with an analogous loss and negative sampling used for subject queries defined in Appendix \ref{analogous gating}.

% Will: I am removing the complication of the object vs. subject queries here. I will add a sentence saying. Note that without loss of generality, we defined the above loss over object queries (as in Section 3.3), with an analogous loss and negative sampling being on on subject queries. 
\cut{
For each triple $\eta = (s, r, o) \in D^{(t)}$, we sample two negative sets of entities $D^-_{\eta, o} = \{o' | (s, r, o') \not\in D^{(t)}\}$ and $D^-_{\eta, s} = \{s' | (s', r, o) \not\in D^{(t)}\}$ and apply the following cross-entropy loss of $L$ is defined as:
\begin{equation}
L_{obj} = - \Big(\sum_{t=1}^T\sum_{\eta \in D^{(t)}}\frac{\exp(\phi_s(s, r, o, t))}{\sum_{o' \in D^-_{\eta, o}} \exp(\phi_s(s, r, o', t))} \\
 + \frac{\exp(\phi_o(s, r, o, t))}{\sum_{s' \in D^-_{\eta, s}} \exp(\phi_o(s', r, o, t))}
\Big)
\end{equation}
}

\begin{table*}[ht]
% \small
 \centering 
 \caption{Temporal KG completion evaluation results on ICEWS, ICEWS05-15 and GDELT. The Hit@1, Hit@3, and Hit@10 metrics are multiplied by 100. Best results are in bold.
  }
 \setlength\tabcolsep{3pt}
  \begin{tabular}{|c|cccc|cccc|cccc|}\hline
     \multirow{2}*{Model}              &\multicolumn{4}{c|}{ICEWS14}   & \multicolumn{4}{c|}{ICEWS05-15}      & \multicolumn{4}{c|}{GDELT}\\
     &\scriptsize{MRR} & \scriptsize{Hit@1} &\scriptsize{Hit@3} &\scriptsize{Hit@10}
     &\scriptsize{MRR} & \scriptsize{Hit@1} &\scriptsize{Hit@3} &\scriptsize{Hit@10}
     &\scriptsize{MRR} & \scriptsize{Hit@1} &\scriptsize{Hit@3} &\scriptsize{Hit@10}\\
     \hline
     TransE  &0.326 &15.4 &43.0 &64.4 &0.330 &15.2 &44.0 &66.0 &0.155 &6.0 &17.8 &33.5  \\
     DistMult  &0.441 &32.5 &49.8 &66.8 &0.457 &33.8 &51.5 &69.1 &0.210 &13.3 &22.4 & 36.5 \\
     ComplEx  &0.442 &40.0 &43.0 &66.4 &0.464 &34.7 &52.4 &69.6 &0.213 &13.3 &22.5 &36.6  \\
     SimplE &0.458 &34.1 &51.6 &68.7 &0.478 &35.9 &53.9 &70.8 &0.206 &12.4 &22.0 &36.6 \\ 
     \hline
     TTransE &0.255 &7.4 &- &60.1 &0.271 &8.4 &- &61.6 &0.115 &0.0 &16.0 &31.8 \\
     HyTE &0.297 &10.8 &41.6 &65.5 &0.316 &11.6 &44.5 &68.1 &0.118 &0.0 &16.5 &32.6 \\
     TA-DistMult &0.477 &36.3 &- &68.6 &0.474 &34.6 &- &72.8 &0.206 &12.4 &21.9 &36.5 \\
    %  DE-TransE &0.465 &29.8 &56.8 &78.0  &0.476 &30.3 &58.3 &80.0 &0.139 &4.6 &16.5 &31.3 \\
    %  DE-DistMult &0.502 &38.0 &56.3 &74.2 &0.619 &49.0 &70.0 &86.4 &0.214 &13.6 &22.6 &35.9 \\
     
     DE-TransE &0.326 &12.4 &46.7 &68.6 &0.314 &10.8 &45.3 &68.5 &0.126 &0.0 &18.1 &35.0 \\
     DE-DistMult &0.501 &39.2 &56.9 &70.8 &0.484 &36.6 &54.6 &71.8 &0.213 &13.0 &22.8 &37.6  \\
     DE-SimplE &0.526 &41.8 &59.2 &72.5 &0.513 &39.2 &57.8 &74.8 &0.230 &14.1 &24.8 &40.3  \\
    %  DE-ComplEx &0.483 &36.7 &54.0 &71.3 &0.583 &45.2 &66.0 &83.5 &0.216 &13.9 &22.9 &36.2 \\
     
     AtiSEE &0.569 &46.3 &63.9 &76.3 &0.520 &39.7 &59.5 &77.3 &- &- &- &-\\
     AtiSER &0.571 &46.5 &64.3 &75.5 &0.484 &35.0 &55.8 &74.9 &- &- &- &- \\
     TNTComplEx &\textbf{0.620} &\textbf{52.0} &66.0 &76.0 &0.670 &\textbf{59.0} &71.0 &81.0 &- &- &- &-\\
     \hline
     TED &0.441 &35.3 &49.1 &60.8 &0.503 &40.8 &56.1 &68.4 &0.237 &14.9 &26.3 &40.7\\
     SRGCN    &0.604 &48.3 &68.0 &83.0 &0.662 &53.5 &74.7 &89.9 &0.239 &15.7 &25.6 &39.8  \\
    %  \hline
     \hline
     \hline
     TeMP-GRU &0.601 &47.8 &68.1 &82.8 &\textbf{0.691} &56.6 &\textbf{78.2} &\textbf{91.7} &\textbf{0.275} &\textbf{19.1} &\textbf{29.7} &\textbf{43.7} \\
     TeMP-SA & 0.607 &48.4 &\textbf{68.4} &\textbf{84.0} &0.680 &55.3 &76.9 &91.3 &0.232 &15.2 &24.5 &37.7 \\
     \hline
     
  \end{tabular} 
  \label{comparative_study}
\end{table*}

\section{Experiments}
We evaluate the performances of TeMP models on three standard TKGC benchmark datasets and analyze the strengths and shortcomings when answering queries with different characteristics. 
Code to reproduce all our experiments is included in the submission and will be made publicly available. 
%We aim to answer the following research questions:

\subsection{Datasets}
We evaluate our model on Global Database of Events, Language and Tone (GDELT) \citep{leetaru2013gdelt} and Integrated Crisis Early Warning System (ICEWS) \citep{DVN/28075_2015} datasets. For ICEWS, we use the two subsets generated by \citet{garcia2018learning}: ICEWS14, corresponding to the facts in 2014 and ICEWS 05-15, containing all facts from 2005 to 2015. For GDELT, we use the subset provided by \citet{trivedi2017know} corresponding to facts from April 1, 2015 to March 31, 2016. We utilize the same partitioning of train, validation and test set as specified by \citet{goel2019diachronic}. More dataset statistics are summarized in Appendix \ref{section dataset statistics}.

\subsection{Evaluation Metrics}
For each quadruple $(s, r, o, t)$ in the test set, we evaluate two queries $(s, r, ?, t)$ and $(?, r, o, t)$. For the first query we calculate scores for $(s', r, o, t), \forall s' \in E$ using Equation \eqref{decoder}. Similar procedure applies to the second query. We then calculate the metrics based on the rank of $(s, r, o, t)$ in each query. Evaluation is performed under filtered settings defined by \citet{bordes2013translating}. 
We report the Hits@1,@3, @10 scores and MRR (mean reciprocal rank).
Please see Appendix \ref{metrics definition} for detailed definitions. 

\subsection{Baseline Methods}
We compare TeMP against a broad spectrum of existing approaches, including a novel rule-based baseline, static embedding methods, and existing state-of-the-art approaches for TKGC. 
%We compare our models against three types of methods and highlight the performance advantage of our methods on TKGC.  The first baseline is a rule-based approach which serves as an indication of how accurate we can answer a query by extracting information from temporal facts alone. The other two types of baselines are graph representation learning methods from existing literature.
% To evaluate the effect of history and future events on predicting occurrence of the current event,

\xhdr{TED model}
We propose a rule-based baseline by directly copying facts from quadruples in the recent past and future, denoted as temporal exponential decay (TED) model. 
The basic idea in this approach is that we predict missing facts by simply copying facts from nearby time steps. The probability of copying each fact is dependent on (1) number of elements overlapping with the queried quadruple and (2) temporal distance to the current time step.
For a detailed description of this baseline, please refer to Appendix \ref{TED formulation}. 
%For more detailed description please refer to Appendix \ref{TED formulation}. 

% \paragraph{Sequence to Sequence}
% We use two Bidirectional GRUs to model the sequence of objects with shared subject-relation pair. At each time step, we use mean aggregator to compute the representation of the set of objects connected to target subject-relation pair. We use 

\xhdr{Static KGC methods} We include TransE \citep{nguyen2016stranse}, DistMult \citep{yang2014embedding}, ComplEx \citep{trouillon2016complex} and SimplE \citep{kazemi2018simple} in the realm of static KG embedding methods. 
We also include a Static RGCN baseline (denoted as SRGCN), which implements the RGCN message-passing approach proposed by \citet{schlichtkrull2018modeling}.
%implemented with TransE, DistMult and ComplEx as decoding function separately. 
Note that all these static baseline methods are employed without considering the time information in the input.

\xhdr{Temporal KGC methods}
We also compare with state-of-the arts models designed for TKGC including TTransE \citep{leblay2018deriving}, TADistMult \citep{garcia2018learning}, HyTe \citep{dasgupta2018hyte}, Diachronic Embedding (DE) \citep{goel2019diachronic}, AtisEE, AtisER \citep{xu2019temporal} and TNTComplEx \cite{lacroix2020tensor}. We don't compare with RE-NET, GHN \citep{han2020graph}, DartNet \cite{garg2020temporal} and Know-Evolve since these work focus on graph extrapolation task.

\subsection{Implementation and Hyperparameters}
All the models except TED are implemented in PyTorch, making use of the PyTorch lightning module and the Deep Graph Library \citep{wang2019deep}.
%We used PyTorch lightning module which supports multi-gpu and multi-node training for model training and experiment management. 
%We implement the RGCN based model directly with the support of Deep Graph Library \citep{wang2019deep}.
We set the negative sampling ratio to 500, i.e. 500 negative samples per positive triple. Because we corrupt subjects and objects separately, there are in total 1000 negative samples collected to estimate the probability of a factual triple. 
For full details on all the model hyperparameters for TeMP and the baselines, refer to Appendix \ref{extra implementation}.

\subsection{Results and Analysis}

% With experimental setup in place, 
% We report comparisons to the prior state of the art, as well as our fine-grained error and ablation analyses. 

\subsubsection{Comparative Study}\label{Q1}

% \todo[inline]{I am waiting for the final experiments to finish before completing table \ref{comparative_study}}

% \todo[inline]{For ICEWS15, GRRGCN results are missing; best model for BiGRRGCN is ensemble only}
% \todo[inline]{For GDELT, the ensemble of GRRGCN and BiGRRGCN are not as good because it takes longer to train and we used smaller batch size}

We compare the baseline models with two instantiations of the TeMP framework: \emph{TeMP-GRU}, \emph{TeMP-SA}, corresponding to the GRU and self-attention variants discussed in Section \ref{sec:temporal}. Incorporating imputation or frequency-based gating is treated optional and we explore different model variants in Section \ref{Q2}. Results on each dataset are given by the model variant that achieves the best validation set performance.
% For simplicity, here we use the name \emph{TeMP-GRU} and \emph{TeMP-SA} without detailed setting.
%For each model, we determine whether adopting unidirectional or bidirectional setting with their performances on validation set. We use bidirectional model for both TeMP-GRU and TeMP-SA models on all three datasets except for one case, where unidirectional TeMP-GRU is used on ICEWS14. 
The core experimental results are summarized in Table \ref{comparative_study}. 

\xhdr{TeMP achieves a new state of the art}
We find that TeMP-SA and TeMP-GRU achieve state-of-the-art results on all three datasets in terms of Hits@10. 
Compared to the most recent work \cite{lacroix2020tensor}---which achieves the best performance to-date on the ICEWS datasets---our results are 8.0\% and 10.7\% higher on the Hits@10 evaluation, though they are slightly worse on Hits@1.  
Additionally, our model achieves a 3.7\% improvement on GDELT compared with DE, the prior state-of-the-art on that dataset. 
The results of the AtiSEE and TNTComplEx methods on the GDELT dataset are not available.

\xhdr{Strong baseline performance}
Interestingly, we find that two of our proposed baseline models also achieve surprisingly strong performance, even outperforming the prior state of the art in some settings.
For example, our rule-based TED baseline achieves relatively strong performance on all three datasets, in particular on GDELT, where it is better than all existing neural models by all measures. 
This highlights the power of simply copying temporal facts with the same patterns as the queried quadruples. 
Similarly, our static RGCN baseline (\emph{SRGCN}) also achieves very strong performance, with the next-best Hits@10 results behind the TeMP framework. 
We hypothesize that the message-passing procedure in SRGCN allows the model to leverage multi-hop structural information that is specific to each time-step, enabling strong performance. 

\cut{
We first notice that SRGCN achieves the performance besides TeMP models in terms of Hit@10 evaluation. This is indicative of the good generalization capability of static message passing network across different time steps. Static RGCN replaces shallow embedding model with message passing network. In this way, the model is capable of capturing multi-hop neighboring information in different KG snapshots. For each entity, static RGCN model learns to aggregate neighborhood edge information in different graph context to produce meaningful embedding, which is essential for link prediction task.
% We compare the relative error reduction rate of TeMP models and baseline models with respect to TED in Table (\ref{REER}). Our proposed methods are the only ones that achieved positive RERR on GDELT dataset.
}

\begin{comment}

\begin{table*}[!h]
 \centering 
 \small
 \setlength\tabcolsep{3pt}
  \begin{tabular}{|c|c|c|}\hline
     \multirow{}               & \multicolumn{1}{c|}{ICEWS05-15}      & \multicolumn{1}{c|}{GDELT}\\
     \hline
     DE-SimplE  & 35.7 & -0.6\\
     AtiSEE  & 42.1 & -\\
     AtiSER  & 36.0 & -\\
     SRGCN-ComplEx  & 74.2 & -0.1\\
     \hline
     GRRGCN-ComplEx  & 76.5 & 4.2\\
     BiGRRGCN-ComplEx  & 78.8 & 4.2\\
     
     \hline
  \end{tabular} 
  \\\caption{Relative Error Reduction Rate(RERR) of hit@10 for different models compared to TED}
  
  \label{REER}
\end{table*}

\end{comment}

\subsubsection{Exploration of Model Variations}\label{Q2}

We study the effect of the imputation and frequency-based gating approaches proposed in Section \ref{sec:heterogeneity} by running model variants on three datasets. We highlight the performance comparison as well as the implication of dataset characteristics on the performance variations. 

Our results are reported on the corresponding validation sets of these benchmarks. 
The results regarding the incorporation of imputation (IM) and frequency-based gating (FG) are shown in Table \ref{ablation-sa}. 
We use a $\checkmark$ to indicate a certain component being used in the experiment, and blank for the absence of the corresponding component. 
\footnote{Imputation is an intrinsic part of TeMP-SA thus it is used in all experiments. See Appendix \ref{sec:extended imputation} for details.}

\xhdr{ICEWS14}
On the ICEWS14 dataset, we find that combining both TeMP-GRU and TeMP-SA models with both imputation and gating achieves the best results on validation set (3.3\% improvement). Additionally, each individual component helps improve the overall model performance by about 1\%.
%Since ICEWS05-15 and GDELT contains significantly more temporal facts than those in ICEWS14, most experiments have not reached convergence before the time limit. We thus compare the validation results of all experiments after training for 39 epochs. 

\xhdr{ICEWS05-15}
On ICEWS05-15, models with gating improved the performance by more than 1\% compared to those without gating. 
However, the additional incorporation of imputation does not result in improvement in the results. 

\xhdr{GDELT}
As for GDELT dataset, we find neither imputation nor gating is significant for model performance. However, it is evident from dataset characteristics that GDELT does not exhibit the same temporal variability and sparsity as the ICEWS datasets. Discussion in Appendix \ref{section dataset statistics} shows that all entities are active at every time step in GDELT (unlike the ICEWS datasets). Additionally, on average each active entity has roughly 150 reference temporal facts in the last 15 time steps, suggesting that each entity involved in TKGC queries are sufficiently informed by the nearby KG snapshots. Data imputation and gating methods are thus unnecessary complexities in GDELT.

% In contrast, since there are significantly more edges involved, more GPU memory space is required with the same experiment setting. This led to the reduction of batch size as well as the number of training epochs completed before the time limit.

% This partially explains the sub-optimal performance compared to other model variants.

% Coupled with temporal recurrence and temporal attention network, we want to study the effect for each individual technique and recommend the best possible combination of these techniques to optimize the task performance. 
% we conducted a subset of $2^6$ experiments on each dataset covering all possible combinations and report the relative performance gain for each of the technique. We show the effect of each technique using statistical test. 

\begin{table}[t!]
\small
 \centering 
   \caption{MRR results for different model variations on ICEWS14, ICEWS05-15 and GDELT
  }
 \setlength\tabcolsep{2pt}
  \begin{tabular}{|c|c|c|c|c|c|}\hline
    {Model} & {IM} & {FG} &{ICEWS14} & {ICEWS05-15} & {GDELT}\\
    \hline
    TeMP-GRU & \checkmark & \checkmark &\textbf{0.610}  &0.680  &0.269 \\ 
    TeMP-GRU &  & \checkmark  &0.599 &\textbf{0.689} & 0.270\\
    TeMP-GRU & \checkmark &  &0.593 &0.670 & \textbf{0.275}\\
    TeMP-GRU &  &  &0.577 &0.673 & 0.274\\
    % TeMP-GRU &  &  &  &0.579 &0.659 \\
    \hline
    TeMP-SA & \checkmark & \checkmark &\textbf{0.623} & \textbf{0.676} & 0.233\\ 
    TeMP-SA & \checkmark &  &0.619 &0.670 & \textbf{0.235}\\
    % TeMP-SA &  & \checkmark &  &0.613 & 0.670\\
    \hline
  \end{tabular} 
  \label{ablation-sa}
\end{table}

\begin{figure*}[!htb]
  \centering
  \includegraphics[width=1\linewidth]{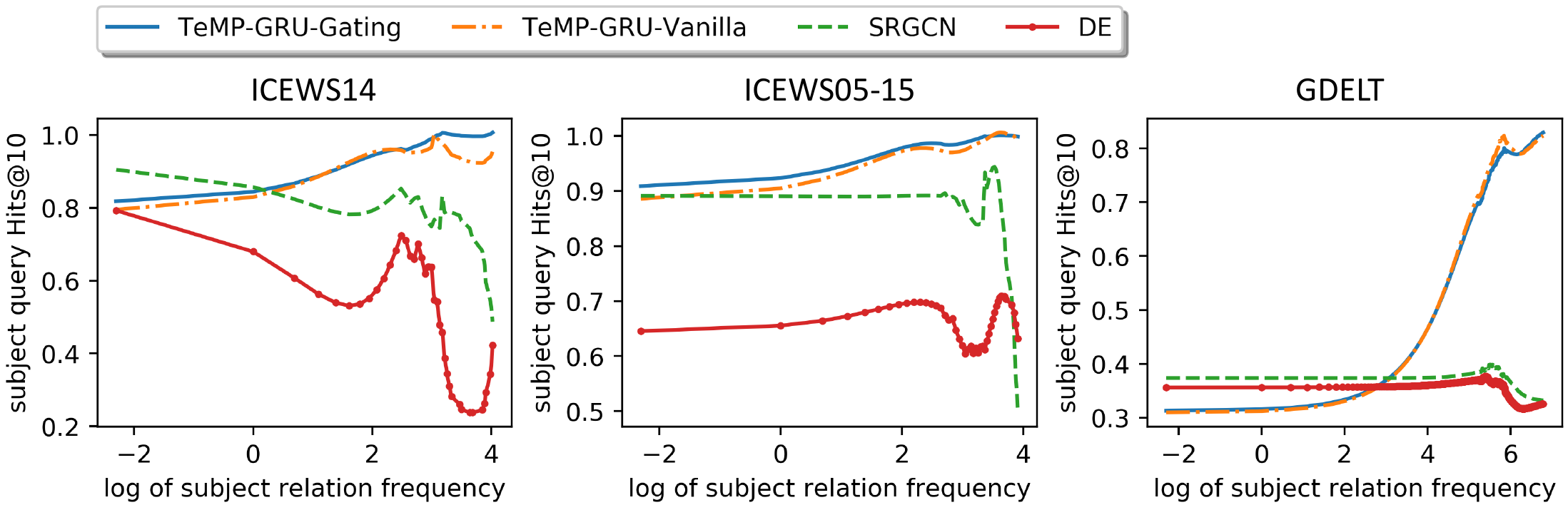}  
\caption{Subject query hit@10 performance comparison of TeMP with different variations and baseline methods.}
\label{subject relation subject}
%   \vspace{<whatever>}
\includegraphics[width=1\linewidth]{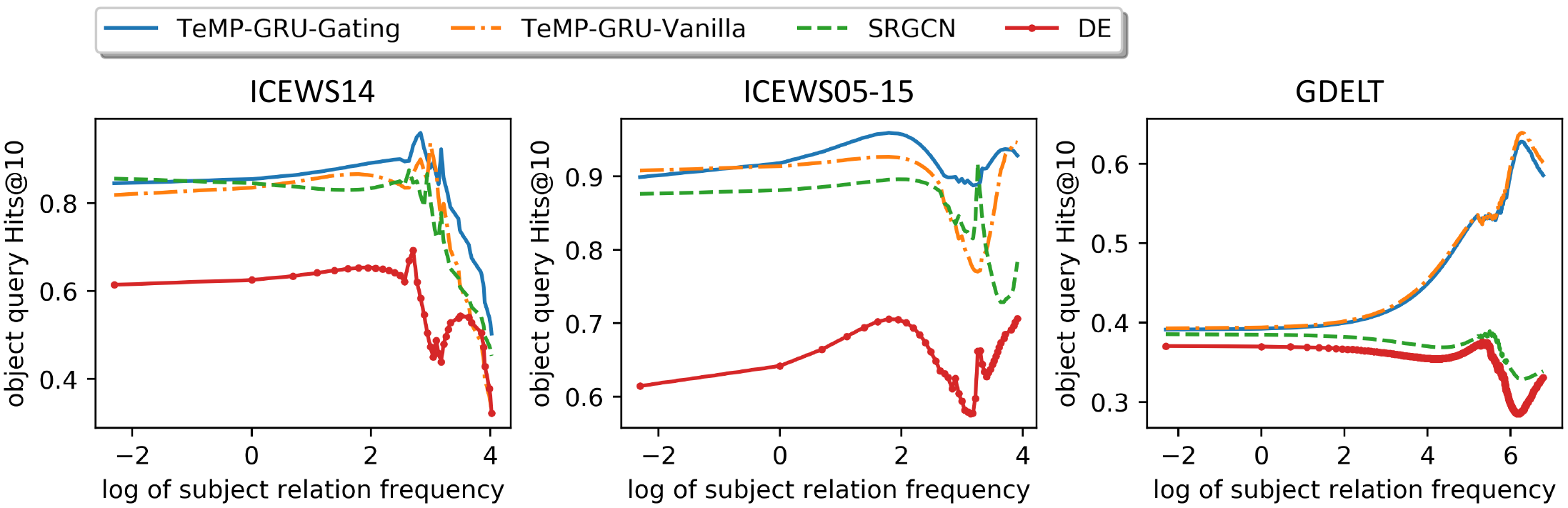} 
\caption{Object query hit@10 performance comparison of TeMP with different variations and baseline methods.}
\label{subject relation object}
\end{figure*}

\subsubsection{Fine-grained Error Analysis}\label{Q3}
To assess how models perform on TKGC queries with different temporal pattern frequencies (TPFs; see Section \ref{sec:heterogeneity}), we group queried quadruples based on different TPFs and calculate the Hits@10 metrics in each group. 
%We omit the original scatter plot and use smoothed curve instead for better clarity.

We plot the temporal subject-relation frequency $f_{s, r}^t$ (defined in Section \ref{sec:heterogeneity}) versus the model performances on subject and object queries to study the \emph{replication} and \emph{reference} effects of temporal facts, respectively. 
Here, we use the term {\em replication effect} to denote the situation where the model can make predictions by copying the exact correct answer to a query from temporal facts. For example, copying \textit{China} from \textit{(Biden, visit, China, 2013)} to answer the query \textit{(Obama, visit, ?, 2014)}.
We use the term {\em reference effect} to denote the effect of having facts that are related (but do not not contain answer entity) to the query fact in the temporal context. For example, selecting \textit{China} from a set countries where Obama visited in the year 2013.

We compare the performances of static models (DE and SRGCN) and temporal models (TeMP-GRU models) on different TPFs. \emph{TeMP-GRU-Vanilla} represents the vanilla version of the model and \emph{TeMP-GRU-Gating} refers to TeMP-GRU model combined with gating technique.  
% \textbf{TeMP-GRU-Ensemble} is a combination of pretrained SRGCN and TeMP-GRU-Vanilla. It uses the weighted combination of logits output by the two pretrained models, where the ensemble weight is calculated based on the same TPF vectors used in frequency-based gating method. 
% The ensemble formulation is similar to equation (\ref{ensemble-1}) - (\ref{ensemble-4}).
Detailed analysis regarding TKGC performance versus other TPFs are discussed in Appendix \ref{anlysis appendix}. 

% \paragraph{Ensemble of Pre-trained Static and Temporal Models}

% To leverage the advantages of SRGCN and TeMP on different temporal frequencies, we explore dynamically combining the two models after training them both. Let \emph{TeMP-Ensemble} denote such model, we use the weighted combination of output logits of SRGCN and TeMP-GRU. The ensemble weight formulation is similar to equation (\ref{ensemble-1}), (\ref{ensemble-2}), (\ref{ensemble-3}) and (\ref{ensemble-4}). We use the temporal frequencies of each queried as features to determine the weights for SRGCN. Results are shown in \ref{ensemble_table}

\paragraph{Replication effect analysis}

% The results and analysis of TED model (section \ref{Q1}) shows that copying subject or object entities from temporal quadruples that share entities and relations with the queried results in moderate result on TKGC. We refer to this as \emph{reference} effect of the temporal facts.

% Inspired by the results and analysis of TED model (section \ref{Q1}), we hypothesize that temporal models tend to answer queries by \emph{copying} object or subject entities from the recent temporal facts sharing elements with the queried quadruple, similar to the defined inference rule of TED (section \ref{TED formulation}). 
%We refer to the copying behavior as \emph{replication} effect of the temporal facts.
% Similar to the defined inference rule of TED (section \ref{TED formulation}), one might expect the temporal model to \emph{copy} the answer "Obama" from \emph{(Obama, visits, Canada, 2012)} when answering the query \emph{(?, visits, China, 2014)}. We refer to this as \emph{reference} effect of the temporal facts.
Here, we examine how the subject-relation TPF correlates with model performance on subject queries.
% A high TPF means that the exact subject-relation pair of a queried quadruple occur frequently in the nearby temporal context.
% (i.e., meaning that the model can predict the correct answer simply by copying the corresponding \textit{subject}). 
Figure \ref{subject relation subject} illustrates that temporal models exhibit positive correlation between subject-relation TPF and subject query performance, while static models show relatively negative correlation between the two quantities. This suggests that the replication effect is stronger in TeMP, indicating that the TeMP model is better at utilizing temporal information for TKGC queries. Additionally, gating helps improve over the vanilla version by a slight margin on all subject-relation frequency values. 
On the other hand, SRGCN achieves better performance on low-TPF queries than temporal models. However, coupling the TeMP model with gating helps close the gap, sometimes surpassing SRGCN on such queries.

\xhdr{Reference effect analysis}
% We also explore how patterns that do not contain the queried entities affect the model performance, e.g., how the number of quadruples such as \emph{(Obama, visit, India, 2012)} affects the results for query \emph{(Obama, visit, ?, 2014}). 
%We name this \emph{reference} effect of temporal facts.
% The visibility of temporal frequencies are the same as defined in section \ref{temporal frequency}. 
% We compare object query results against (1) and (4), then compare subject query results against (2) and (5).
Here, we examine how the occurrence of related facts (not containing the answer) in the temporal context impacts performance. 
We find that the temporal models exhibit non-linear correlations between object query performance and subject-relation TPF (Figure \ref{subject relation object}). 
In particular, on the ICEWS datasets the performance increases as the log-frequencies grows from $-\infty$ to 2 and drops at higher frequency values. 
%Compared to the near-constant positive correlation in Figure \ref{subject relation subject}, 
%We extrapolate that the increase in the number of temporal facts \emph{(Obama, visit, *, $t'$)}, $t' \in \{2008, ... 2020\}$ makes it more difficult for temporal models to choose the right answer from the set of countries associate with \emph{Obama} and \emph{visit}. 
% this indicates that temporal models may be confused about high density information in neighboring KG snapshots.
We hypothesize that it is harder for temporal model to select the answer from a very large set of object candidates, e.g,. choosing \textit{China} from more than 100 countries that Obama visited from 2008 to 2013.
In terms of model comparisons, we find that gating helps TeMP-GRU to surpass its vanilla version and SRGCN on most TPF values. The margin of improvement is especially significant on queries of high TPF in ICEWS05-15.
% SRGCN shows little performance variations at low frequency values and decrease sharply at higher frequency.

The null effect of frequency-based gating on GDELT can be attributed to the same reason as discussed in Section \ref{Q2}.
\begin{comment}

\begin{table*}[h]
 \centering 
   \caption{Results of TeMP-Ensemble on ICEWS14, ICEWS05-15 and GDELT}
  \label{ensemble_table}
 \setlength\tabcolsep{3pt}
  \begin{tabular}{|c|cccc|}\hline
  Dataset & MRR & Hit@1 &Hit@3 &Hit@10\\
     \hline
     ICEWS-14 &0.601 &46.5 &69.4 &84.7 \\
     ICEWS05-15 &0.711 &59.2 &79.6 &92.4 \\
     GDELT &0.264 &17.9 &28.5 &43.0 \\
     \hline
     
  \end{tabular} 
\end{table*}

\end{comment}

\section{Conclusion}
In this work, we present a novel framework named TeMP for temporal knowledge graph completion (TKGC). TeMP computes entity representation by jointly modelling multi-hop structural information and temporal facts from nearby time-steps. 

% \red{It is a generic framework where both the structural and temporal encoder are not limited to the model architecture we used in the paper.}

Additionally, we introduce novel frequency-based gating and data imputation techniques to address the temporal variability and sparsity problems in TKGC. We show that our model is able to achieve superior performance (10.7\% relative improvement) over the state-of-the-arts on three benchmark datasets. Our work is potentially beneficial to other tasks such as temporal information extraction and temporal question answering, by providing beliefs about the likelihood of facts at particular points in time.

Future work involves exploring the generalization of TeMP to continuous TKGC and better imputation techniques to induce representations for infrequent and inactive entities.
\paragraph{Acknowledgements}
This research is supported by CIFAR Canada AI Chair program and Samsung Electronics. The authors would like to thank Compute Canada for providing the computational resources. 
\bibliography{emnlp2020}
\bibliographystyle{emnlp2020}
\appendix

\newpage
\pagenumbering{arabic}
\setcounter{page}{1}
\section{Appendix}

\subsection{Architecture Details}
\xhdr{Temporal Edge Dropout}\label{freq dropout}
% The temporal variability problem enables temporal model to copy from facts from nearby KG snapshot, we refer to this behavior as "overfitting" to the historical facts. In order to alleviate the overfitting problem, we propose historical edge dropout: randomly dropping historical facts in the graph evolution process used to induce the current entity representation.
% The access to temporal facts in nearby KG snapshots enables temporal model to \emph{copy} subject or object entities, we refer to this behavior as

The replication effect illustrated in Figure \ref{subject relation subject} and \ref{replication figures} shows that TeMP is increasingly better capable at copying from temporal facts when TPFs also increase. We refer to this as "overfitting" to the temporal facts. In order to alleviate such problem, we propose temporal edge dropout: randomly dropping facts occurred in the defined time window used to induce the entity representation. 

\citet{rong2019dropedge} propose dropping a proportion in the local graph context to combat over-fitting and over-smoothing. We extend this technique to TKG by either (1) randomly dropping a certain percentage of quadruples in each temporal snapshot and (2) drop quadruples with different probabilities based on certain quadruple characteristics. Details of the second method is omitted since we find the two methods working equally well. We use 0.2 as temporal edge dropout rate in all experiments.

\xhdr{Positional Embedding}
We capture the time-sensitive information in the TKG by combining the entity representation with positional embedding. The positional embedding is denoted as $\{\bm{p}_1, \bm{p}_2, ..., \bm{p}_T\}$, which embeds absolute positional information of each time step. The set of representations for entity $e_i$ at all time steps is $\{\bm{p}_1 + \bm{z}_{i, 1}, \bm{p}_2 + \bm{z}_{i, 2}, ..., \bm{p}_T + \bm{z}_{i, T}\}$, which are used as input entity representation to the decoding function.

\subsection{Extended Imputation Formulation}\label{sec:extended imputation}

For bidirectional temporal recurrent model, we defined the imputed representation analogous to Equation \eqref{equation:imputation_1} and \ref{equation:imputation_2}.
We use $t^+$ to denote the very next time step at which entity $e_i$ is active after $t$. The decay rate for imputing from future representations as follows:
\begin{equation*}
\gamma_{i,t^+}^x = \exp\{-\max(0, \lambda_x|t-t^+| + b_x)\}.
\end{equation*}
To calculate the imputed representation of the $e_i$ at time $t$, we divide both exponential decay rates by two and renormalize:
\begin{align*}
\gamma_{i,t}^x &= 1 - \frac{\gamma_{i,t^-}^x}{2} - \frac{\gamma_{i,t^+}^x}{2}\\
\boldsymbol{\hat{x}}_{i, t} &= \frac{\gamma_{i,t^-}^x}{2} \boldsymbol{x}_{i, t^-} + \frac{\gamma_{i,t^+}^x}{2} \boldsymbol{x}_{i, t^+} + \gamma_{i,t}^x \boldsymbol{x}_{i, t} \ .
\end{align*}

\xhdr{Intrinsic imputation for TeMP-SA}
% \xhdr{Positional Embedding}
We use Equation \eqref{equation:attention_1} - \eqref{equation:attention_4} to derive entity representations for both active and inactive entities and view it as an intrinsic way of imputation. Hence imputation is tagged with all TeMP-SA results in Table \ref{ablation-sa}.

\subsection{Analogous Definition of Frequency Based Gating and Training Loss}\label{analogous gating}
We define the process for deriving entity representation for subject queries $(?, r, o, t)$ analogous to Equation \eqref{eq:gating-1} and \eqref{eq:gating-2}. The model is only allowed the access to frequencies $F_o = [f_o^t, f_r^t, f_{o, r}^t]$, we use it to define a similar gating over static and temporal entity representations:
\begin{align*}
\bm{z}_{s, t} &\coloneqq  \alpha_{ss} \bm{x}_{s, t} + (1 - \alpha_{ss}) \bm{z}_{s, t}\\
\bm{z}_{o, t} &\coloneqq  \alpha_{so} \bm{x}_{o, t} + (1 - \alpha_{so}) \bm{z}_{o, t},
\end{align*}
where $\alpha_{ss} = \text{MLP}_{ss}(F_o)$, $\alpha_{so} = \text{MLP}_{ss}(F_o)$,  $\alpha_{ss}, \alpha_{so}  \in [0,1]$. 
With the negative subject entity set being $D^-_{\eta, s} = \{s' | (s', r, o) \not\in D^{(t)}\}$, the training loss for subject queries is defined as follows:
\begin{equation*}
 L_{sub} = - \sum_{t=1}^T\sum_{\eta \in D^{(t)}}\frac{\exp(\phi(s, r, o, t))}{\sum_{s' \in D^-_{\eta, s}} \exp(\phi(s', r, o, t))}.
\end{equation*}

The final training loss is the sum of losses for two types of queries: $ L = L_{sub} + L_{obj}$.

\subsection{Detailed TED Formulation and Analysis}\label{TED formulation}

\xhdr{TED Model Definition}
We hypothesize that certain quadruples with \emph{more frequent} occurrence in \emph{more recent} time steps are informative for the current-step KGC. 
For each query, we construct a set of reference entities from training data. Similar to the down-weighting mechanism of temporal encoder (Section \ref{sec:temporal}), we score each entity based on exponential decaying mechanism with respect to the temporal distance to the current time step. We then rank the entities in the reference set according to such scores.

For each queried quadruple $(s, r, o, t)$, we collect reference entity sets consisting of tuples $\{(e, t'), t' \neq t\}$ where $e$ is the subject or object entity and $t'$ is the corresponding time of occurrence. The tuples are extracted from the temporal facts sharing at least one element with $(s, r, o, t)$. We divide them into subject and object reference sets two types of queries. The subject reference set consists of:
 \begin{itemize}
     \item[(1)] subjects with shared relation-object pair, i.e., $\{(s', t') | \exists t' \neq t , (s', r, o) \in D^{(t')}_{train}\}$,
     \item[(2)] subjects with shared object, i.e., $\{(s', t') | \exists t' \neq t \land r' \in R, (s', r', o) \in D^{(t')}_{train} \}$,
     \item[(3)] subjects with shared relation, i.e., $\{(s', t') | \exists t' \neq t \land o' \in E ,  (s', r, o') \in D^{(t')}_{train}\}$.
 \end{itemize}

Symmetrically, object reference set consists of:

 \begin{itemize}
 \item[(1)] objects with shared subject-relation pair, i.e., $\{(o', t') | \exists t' \neq t , (s, r, o') \in D^{(t')}_{train}\}$,
 \item[(2)] objects with shared subject, i.e., $\{(o', t')  | \exists t' \neq t  \land r' \in R, (s, r', o') \in D^{(t')}_{train}\}$,
 \item[(3)] objects with shared relation, i.e., $\{(o', t') | \exists t' \neq t \land s' \in E, (s', r, o') \in D^{(t')}_{train}\}$.
 \end{itemize}
 
We don't collect triples in the current time step $t$ as we assume $D_{train}^{(t)} \cap D_{test}^{(t)} = \emptyset, \forall t$.

Note that (1) is a subset of (2) and (3), also (2) and (3) contain overlapping tuples. We define the priority to be (1) $>$ (2) $>$ (3), such that if some tuple is present in (1), then it will be removed from both (2) and (3). This is based on the assumption that objects with the same subject-relation pair as the current triple are the most ideal candidates. For example, because of the characteristics of \textit{police}, the fact \textit{(police, arrest, citizen)} occurred multiple times across in the dataset. Objects with same shared subject and different relation comes second, e.g. \emph{(Obama, visit, China, 2013)}, \emph{(Obama, visit, Russia, 2014)} are important information for predicting \emph{(Obama, make\_announcement\_to, ?, 2015)}.

% Let $S$ be some set of tuple defined above, we apply score function $\exp(-\sigma|t - t'|)$ to each tuple $(e, t') \in S$. The score for $e$ is the sum over all tuples containing $e$, i.e.,
Let $S$ be some set of tuple defined above. The score for $e$ is the sum over all tuples containing $e$, 

\begin{equation}
\sum_{t', (e, t') \in S}\exp(-\sigma|t - t'|), \sigma > 0.
\end{equation}
\xhdr{TED Results and Analysis}
Table \ref{Ted model} shows the sensitivity analysis for parameter $\sigma$ on validation set. We notice that the performances are low when $\sigma$ is either extremely large or small, while peaks when $\sigma=0.1$ on ICEWS datasets and $\sigma=1$ on GDELT dataset. This suggests an existing trade-off between \emph{recency} and \emph{frequency} heuristics.

TED model results also expose the bias of recurring events in political event datasets, particularly in GDELT. However, TED should be considered by future work as an important baseline to gauge the relative model performance. Additionally, the results suggests the potential for pointer-style TKGC -- deciding between coping an entity from historical facts and selecting an entity in the current snapshot to answer a query.

\begin{table*}[!htb]
 \centering 
%  \small
 \setlength\tabcolsep{3pt}
  \begin{tabular}{|c|cccc|cccc|cccc|}\hline
     \multirow{2}*{$\sigma$}              &\multicolumn{4}{c|}{ICEWS14}   & \multicolumn{4}{c|}{ICEWS05-15}      & \multicolumn{4}{c|}{GDELT}\\
     &\scriptsize{MRR} & \scriptsize{Hit@1} &\scriptsize{Hit@3} &\scriptsize{Hit@10}
     &\scriptsize{MRR} & \scriptsize{Hit@1} &\scriptsize{Hit@3} &\scriptsize{Hit@10}
     &\scriptsize{MRR} & \scriptsize{Hit@1} &\scriptsize{Hit@3} &\scriptsize{Hit@10}\\
     \hline
     $10^{-5}$  &0.434 &36.2 &48.9 &60.5 &0.466 &36.2 &52.4 &66.6 &0.179 &10.1 &18.7 &33.1   \\
     $10^{-2}$  &0.445 &35.5 &49.9 &61.2 &0.498 &39.7 &55.9 &\textbf{68.8} &0.192 &11.0 &20.3 &35.6     \\
     $10^{-1}$  &\textbf{0.455} &\textbf{36.7} &\textbf{50.7} &\textbf{61.6} &\textbf{0.505} &\textbf{40.8} &\textbf{56.3} &68.7 &0.226 &13.7 &24.7 &40.4     \\
     $1$  &0.449 &35.9 &50.3 &61.4 &0.500 &40.1 &55.9 &68.5 &\textbf{0.238} &\textbf{15.0} &\textbf{26.3} &\textbf{40.8}     \\
     $10^{1}$  &0.449 &35.9 &50.3 &61.5 &0.496 &39.8 &55.4 &68.1 &0.237 &14.9 &26.2 &40.7     \\
     $10^{2}$ &0.446 &35.5 &50.0 &61.2 &0.482 &38.3 &53.9 &66.8 &0.232 &14.4 &25.8 &40.2     \\
     $10^{5}$  &0.359 &24.9 &41.7 &57.2 &0.362 &23.8 &42.8 &60.6 &0.091 &3.0 &8.0 &20.2     \\
     \hline
  \end{tabular} 
  \\\caption{TKGC evaluation results(filtered setting) using TED model under various $\sigma$ values. The Hit@1, Hit@3, and Hit@10 metrics are multiplied by 100. }\label{Ted model}
\end{table*}

\subsection{Dataset Statistics and Characteristics}\label{section dataset statistics}
\begin{figure*}
\begin{subfigure}{.32\textwidth}
  \centering
  % include third image
  \includegraphics[width=1\linewidth]{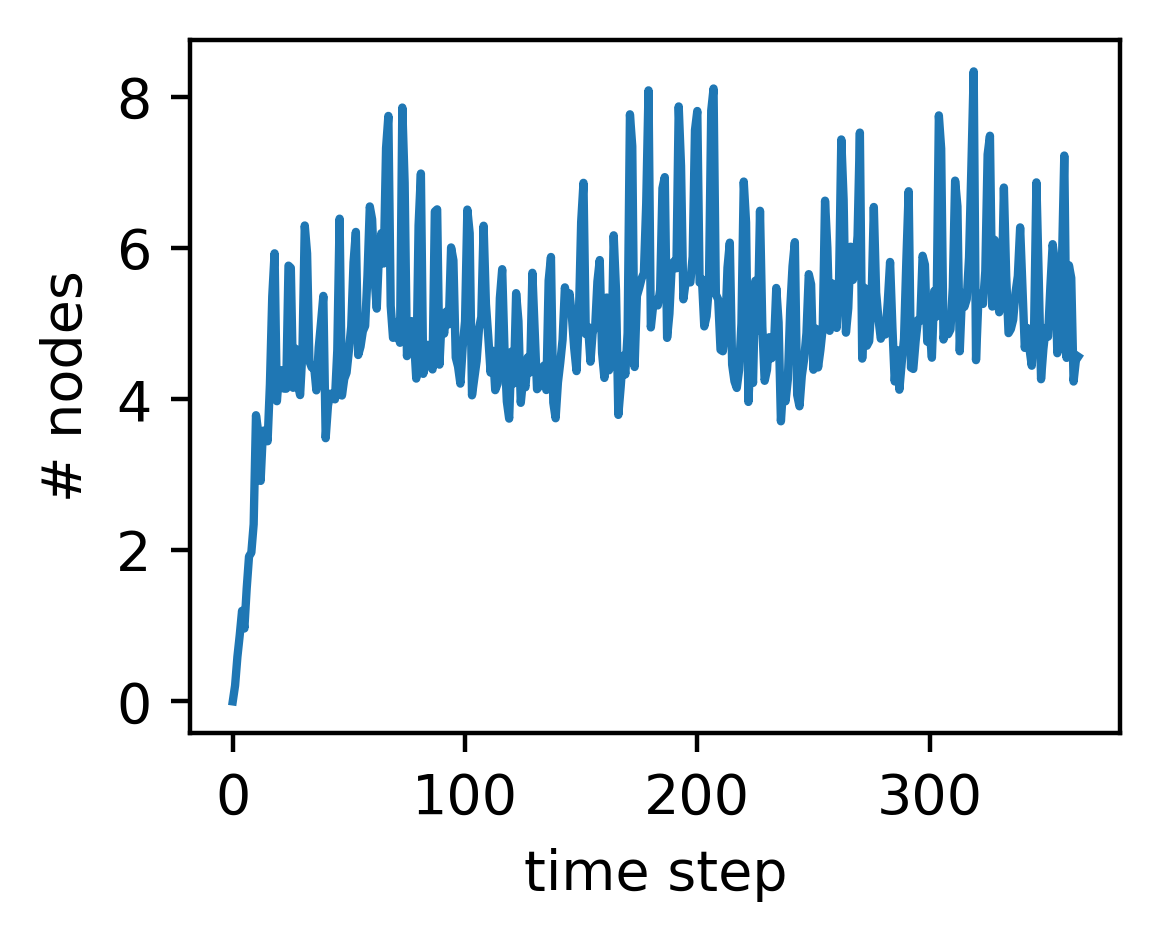}  
  \caption{}
\end{subfigure}
\begin{subfigure}{.34\textwidth}
  \centering
  % include third image
  \includegraphics[width=1\linewidth]{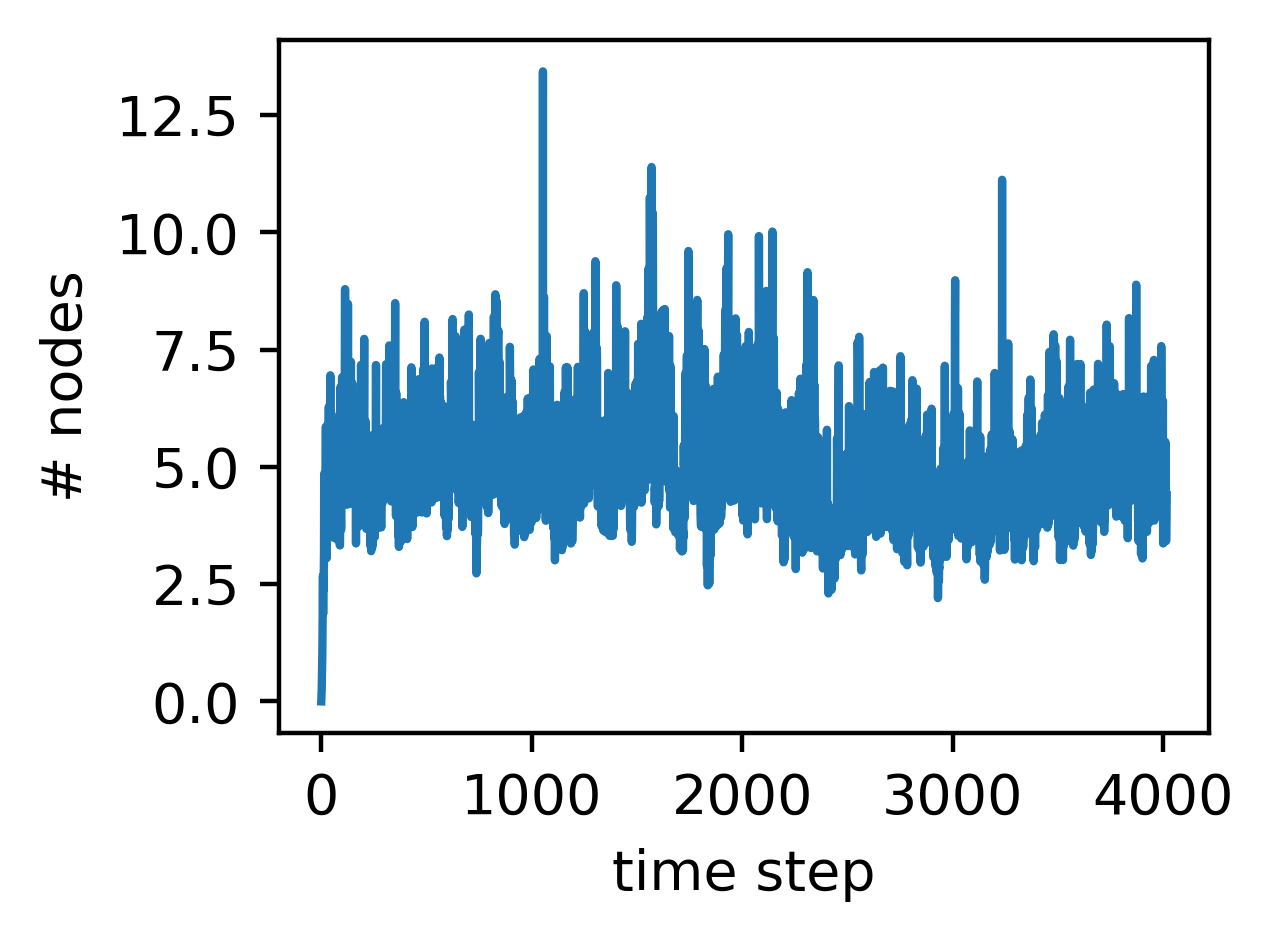}  
  \caption{}
\end{subfigure}
\begin{subfigure}{.33\textwidth}
  \centering
  % include fourth image
  \includegraphics[width=1\linewidth]{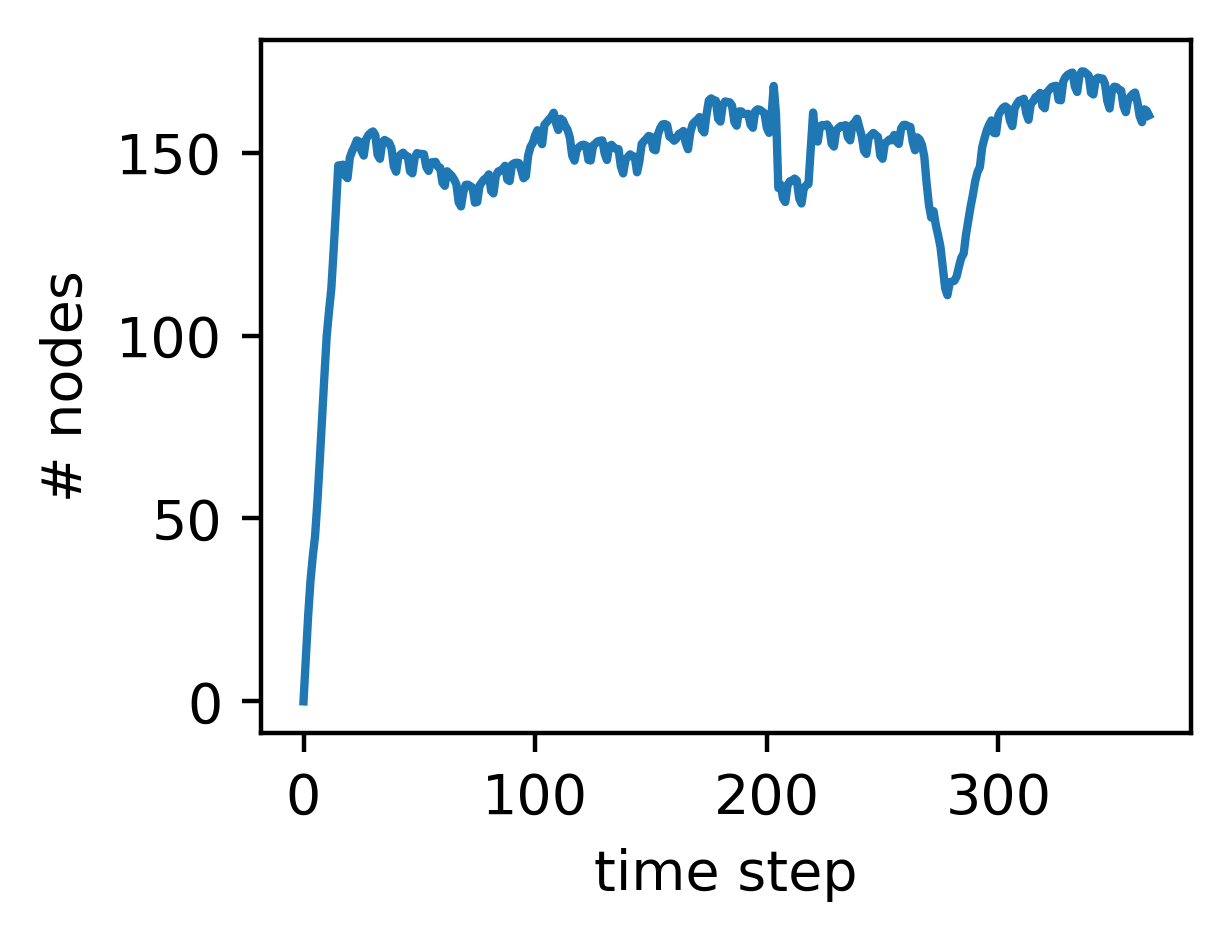}  
  \caption{}
\end{subfigure}
\caption{At each time step, for every active entity we calculate how many times each active entity occurred in that last 15 time steps and take average. We show the distirbution of such quantities on (a)ICEWS14, (b) ICEWS05-15 and (c) GDELT}
\label{num_avg_active_hist_nodes}
\end{figure*}

The dataset statistics are summarized in Table \ref{dataset_statistics}. The numbers of entities are 7,128, 10,488 and 500 respectively in three datasets, indicating that temporal sparsity issue is severe on ICEWS datasets but trivial on GDELT dataset. The temporal variability of three datasets is demonstrated in Figure \ref{num_avg_active_hist_nodes}. The average number of associated temporal facts for each entity is much lower in ICEWS datasets compared to GDELT. The difference can be attributed to the fact that GDELT dataset is constructed by extracting facts among the most frequent 500 entities in the entire dataset. This intrinsically eliminates the sparsity and variability bias in the original datasets.
% The volatility of such quantities across different time steps is larger.

% However it justifies the usage of temporal model, even for ICEWS datasets there are on average 5 temporal facts to be leveraged to inform the current entity representation.

\begin{table*}[!htb]
 \centering
  \begin{tabular}{|c|c|c|c|c|c|c|c|}\hline
     Dataset &\# entities & \# relations &\# time steps &\multicolumn{1}{c|}{N_{train}}   & \multicolumn{1}{c|}{N_{valid}} & \multicolumn{1}{c|}{N_{test}} & \multicolumn{1}{c|}{N_{total}}\\
     \hline
     ICEWS14 & 7,128 &230 &365 &72,826 &8,941 &8,963 &90,730 \\
     ICEWS05-15 & 10,488 & 251 & 4017 & 386,962 & 46,275 & 46,092 & 479,329 \\
     GDELT & 500 & 20 & 366 & 2,735,685 & 341,961 & 341,961 &3,419,607 \\
     \hline
     
  \end{tabular} 
  \\\caption{Statistics of ICEWS14, ICEWS05-15 and GDELT datasets. }
  \label{dataset_statistics}
\end{table*}
\begin{figure*}[!htb]
\begin{subfigure}{.5\textwidth}
  \centering
  % include third image
  \includegraphics[width=.8\linewidth]{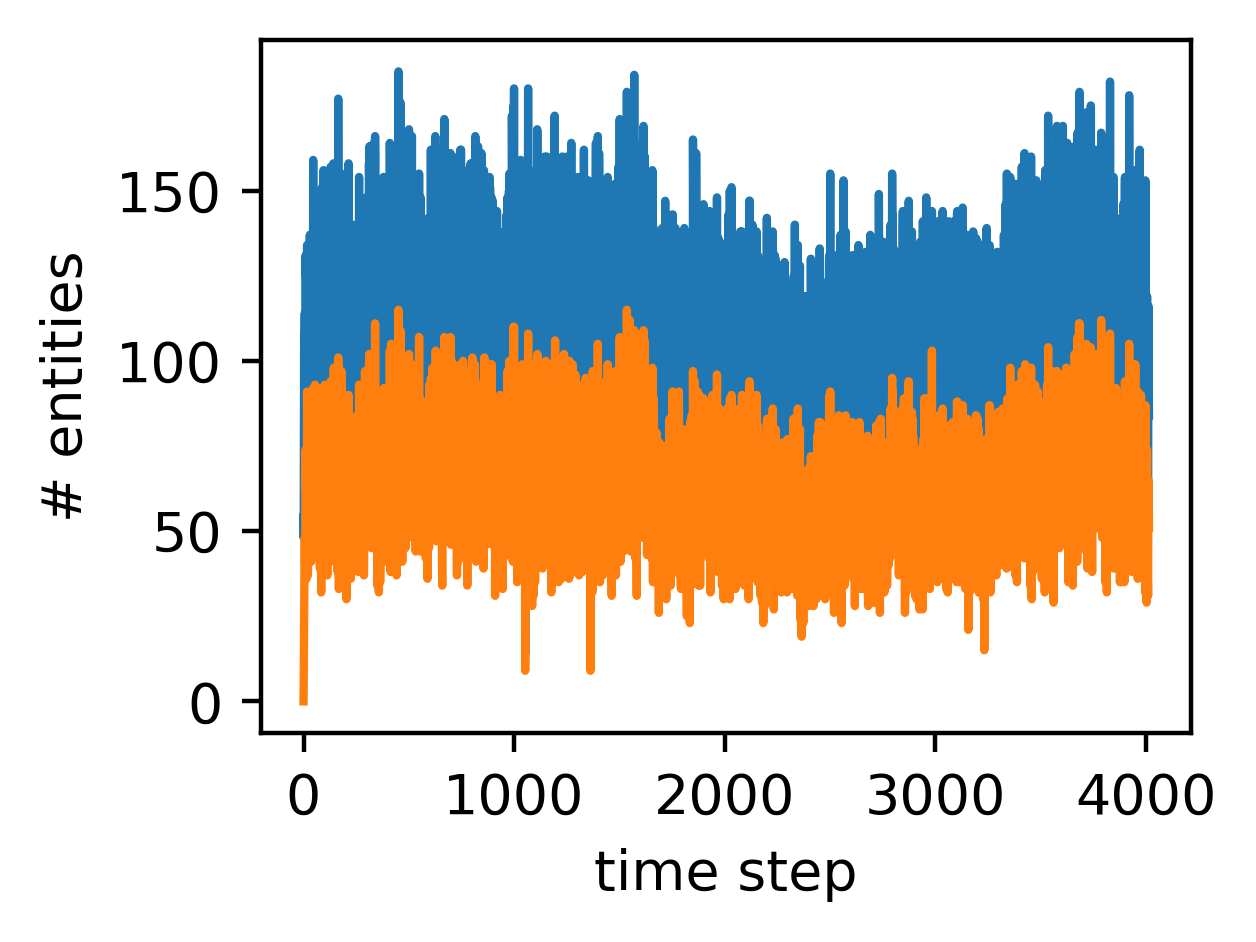}  
\end{subfigure}
\begin{subfigure}{.5\textwidth}
  \centering
  % include fourth image
  \includegraphics[width=.8\linewidth]{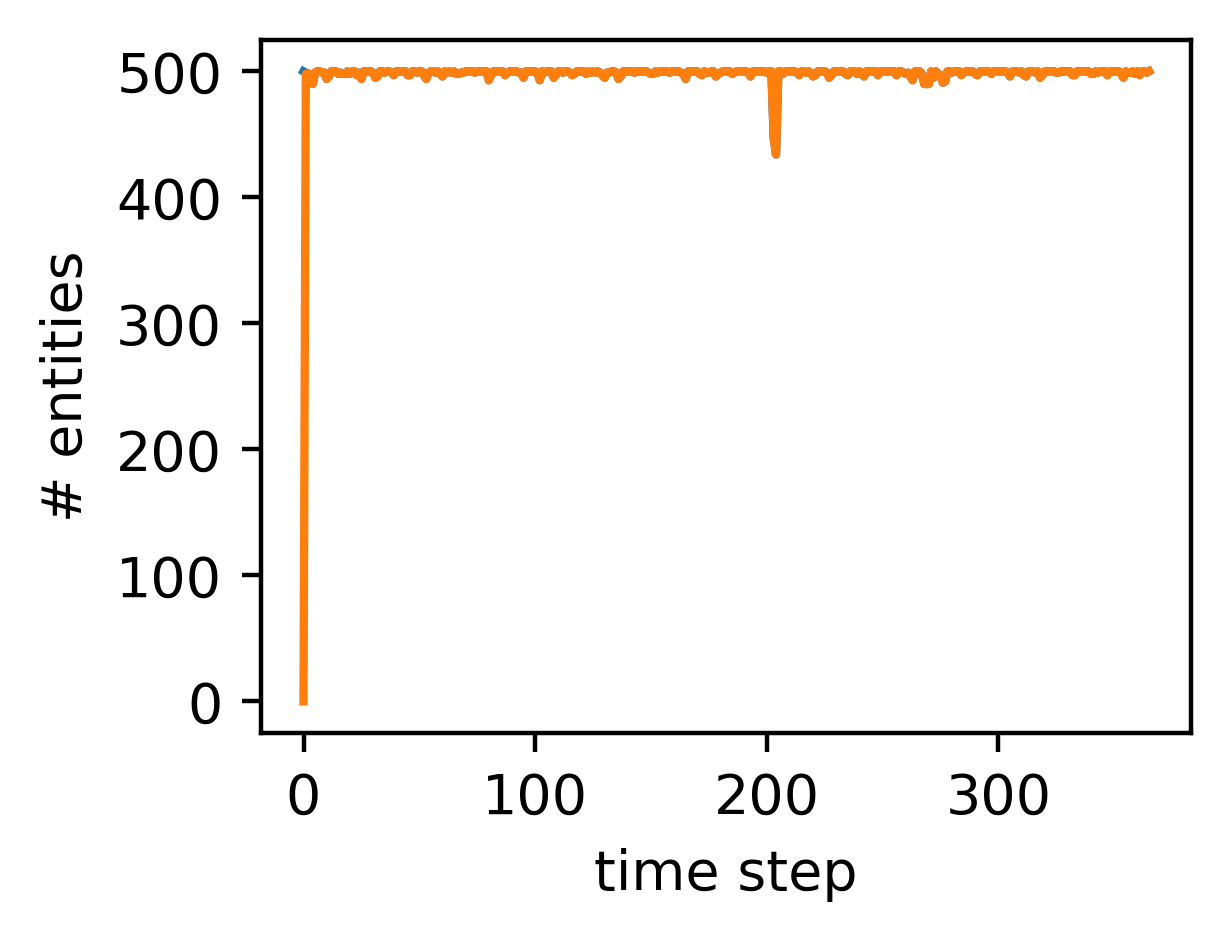}  
\end{subfigure}
\caption{Dataset statistics of ICEWS05-15 (left) and GDELT (right) as a supplement of Figure \ref{num_existing_hist_entities}. }
\label{num_existing_hist_entities_appendix}
\end{figure*}

\subsection{Definitions for Evaluation Metrics}\label{metrics definition}
We use MRR, Hits@1, Hits@3 and Hits@10 to evaluate the model performance. MRR is defined as:
\begin{multline}
\frac{1}{2 * |D_{test}|}\sum_{t=1}^T \sum_{\eta = (s, r, o) \in D_{test}^{(t)}} (\\\frac{1}{\text{rank}(o | s, r, t)} + \frac{1}{\text{rank}(s | r, o, t)})
\end{multline}
The Hit@1, Hit@3, Hit@10 are the percentages of test facts for which the $k$ highest ranked predictions contain the correct prediction, $k = 1, 3, 10$. That is, 
\begin{multline}
\frac{1}{2 * |D_{test}|}\sum_{t=1}^T \sum_{\eta = (s, r, o) \in D_{test}^{(t)}}\\\big(I(\text{rank}(o | s, r, t) \leq k) + I(\text{rank}(s | r, o, t) \leq k)\big)
\end{multline}
where $k = 1, 3, 10$, $I$ is the indicator function.

\subsection{Detailed Implementation and Hyperparameters}\label{extra implementation}
We use the Adam optimizer and set the learning rate to 0.001. 
The batch size is set to 8 for ICEWS14 and ICEWS05-15, i.e. each batch contains facts in 8 snapshots. We additionally sample 3,000 quadruples in each snapshot to avoid out-of-memory issue.\\ 
Embedding size and hidden sizes for both recurrent and self-attentive models are both set to 128. We use 8 attention heads in TeMP-SA to model the multi-faced evolution of TKG. As required by reproducibility checklist, the complete hyperparameter setting and run-time information for TeMP-GRU model on all benchmark datasets are summarized in Table \ref{hyperparameters}.

Suggested by ablation study in \citep{jin2019recurrent} we set the number of relational convolution layers to 2 to encode two-hop neighbors. We apply temporal edge dropout technique to TKG, in each training epoch we randomly drop 50\% of the quadruples in current KG and 20\% triples in each temporal reference KG to combat over-fitting and over-smoothing. We experimented with TransE, DistMult and ComplEx on validation set and found that ComplEx \citep{trouillon2016complex} yields the best performance overall. Hence ComplEx is used as decoding function to score head or tail entities given queries. During inference on $D^{(t)}_{valid}$ and $D^{(t)}_{test}$, our models take $D^{(t-\tau)}_{train},..., D^{(t)}_{train}$ as input and compute the scores to compute the entity representations.

The parameter $\tau$ stands for the number of KG snapshots available for answering query. This is applied to temporal models as a \textit{budget}. Single-direction models take temporal entity embedding from the past $\tau$ graphs while bidirectional models focus on $\frac{\tau}{2}$ historical and future snapshots. 
% Temporal KG snapshots are encoded by both static and temporal relational encoder, the entity representations are used as input for the current-step KGC. 

\begin{table*}[t!]
\caption{Hyperparameters setting for TeMP-GRU model on three benchmark datasets}
\small
 \centering 
 \setlength\tabcolsep{2pt}
  \begin{tabular}{|c|c|c|c|c|c|c|c|}\hline
    {Dataset} & {batch size} & \# temporal snapshots & GPU type &\# GPU &Time limit & runtime per epoch  & \# parameters\\
    \hline
    ICEWS14 & 8 & 15 & GeForce GTX TiTan & 1 & 24h & 8m &885K\\
    \hline
    ICEWS05-15 & 8 & 10 & Nvidia V100 &1 & 60h & 70m &2856K\\
    \hline
    GDELT & 4 & 15 & Nvidia V100 &2 & 60h & 13m &878K\\
    \hline
  \end{tabular} 
  \label{hyperparameters}
\end{table*}

We use early stopping with patience 10 with respect to the average MRR on the validation set. All ablation studies are conducted on the validation set. For the best model variants, we use the model checkpoint that achieves the best MRR score on validation set to perform final evaluation on test set. 

\subsection{Detailed Analysis of Performances versus TPFs}\label{anlysis appendix}
We studied the correlation between subject-relation TPF and query answering performances in Section \ref{Q3}. Here, we first define a complete set of TPFs that covers all possible subsets of a quadruple. In Section \ref{sec:heterogeneity} we defined (1) subject frequency $f_s^t$, (2) object frequency $f_o^t$, (3) relation frequency $f_r^t$, (4) subject-relation frequency $f_{s,r}^t$, (5) relation-object frequency $f_{r,o}^t$ related to quadruple $(s, r, o, t)$. We additionally define (6) subject-object frequency $f_{s,o}^t$ and (7) triple frequency $f_{s,r,o}^t$. We use the following combinations of TPFs and query types to study replication and reference effects respectively. 

\begin{figure*}[!htb]

\begin{subfigure}{1\textwidth}
\includegraphics[width=1\linewidth]{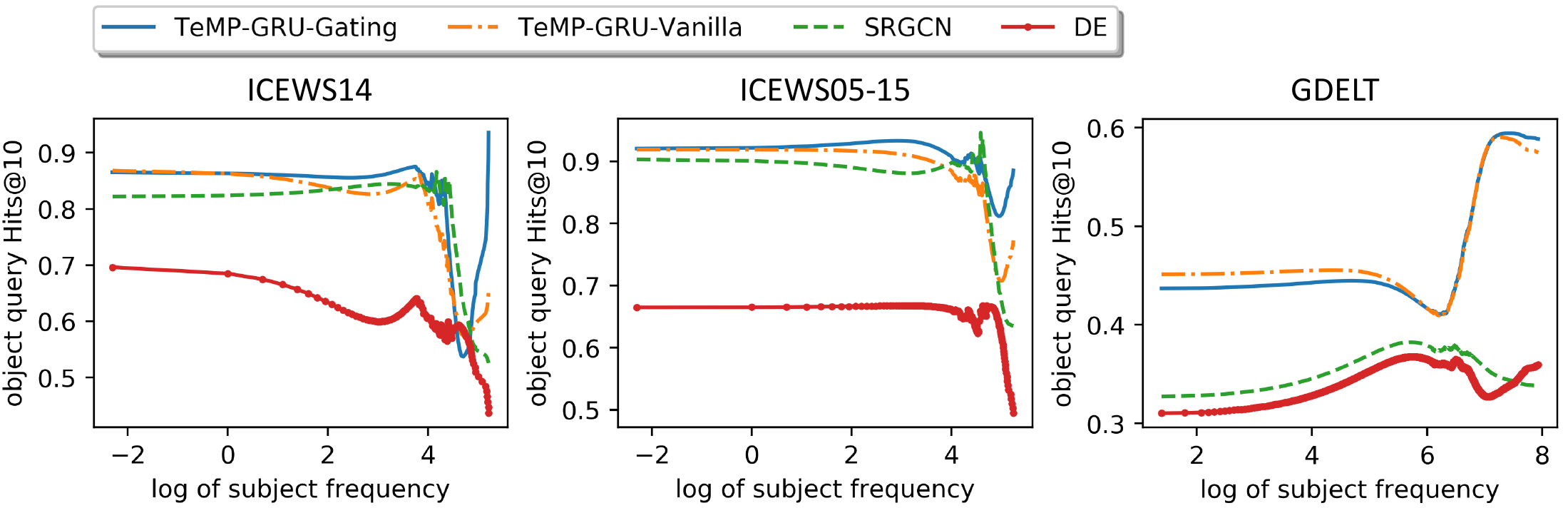} 
\caption{Object query Hits@10 performances versus temporal subject frequencies}
\end{subfigure}
\begin{subfigure}{1\textwidth}
\includegraphics[width=1\linewidth]{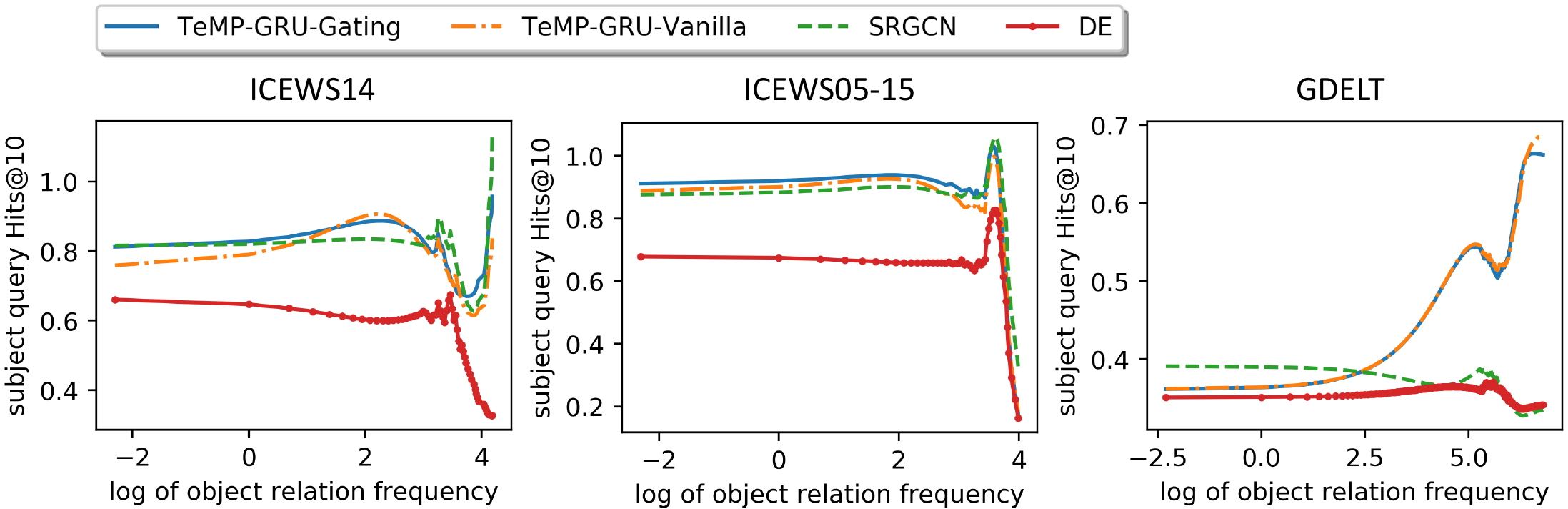} 
\caption{Subject query Hits@10 performances versus temporal relation-object frequencies}
\end{subfigure}
\begin{subfigure}{1\textwidth}
\includegraphics[width=1\linewidth]{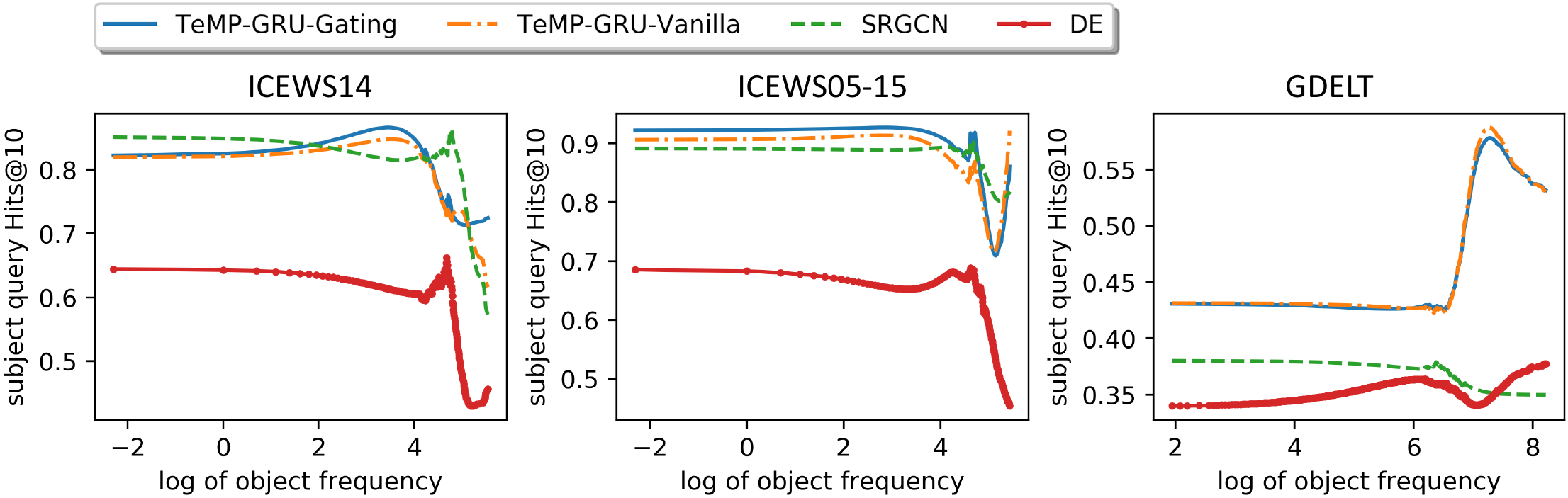} 
\caption{Subject query Hits@10 performances versus temporal object frequencies}
\end{subfigure}

\caption{Plots of reference effect group}
\label{reference figures}
\end{figure*}

For replication effect, we compare subject query results against (1), then compare object query results against (2) and (5). Values of (6) and (7) are compared with the results of both subject and object queries. 
For reference effect, we compare object query results against (1), and subject query results against (2) and (5). 
Results are summarized in Figure \ref{replication figures} and Figure \ref{reference figures} respectively. 

The general observation is similar to the discussion in Section \ref{Q3}. In the replication analysis , TeMP-GRU models show significantly more positive trends than the static models (SRGCN and DE). However, we witness drops in performances when TPFs become large in the reference effect analysis. Performance of TeMP-GRU-Vanilla model improves with the help of gating on ICEWS datasets on TPFs. The benefit is less obvious on GDELT dataset due to the observation that GDELT is less affected by temporal sparsity and variability problem (Appendix \ref{section dataset statistics}).

We conclude that TeMP models are significant more advantageous in utilizing temporal facts for TKGC task. In addition, frequency-based gating improves the overall performance with respect to all different TFPs.

\begin{figure*}[!htb]
  \centering
    \begin{subfigure}{1\textwidth}
  \includegraphics[width=1\linewidth]{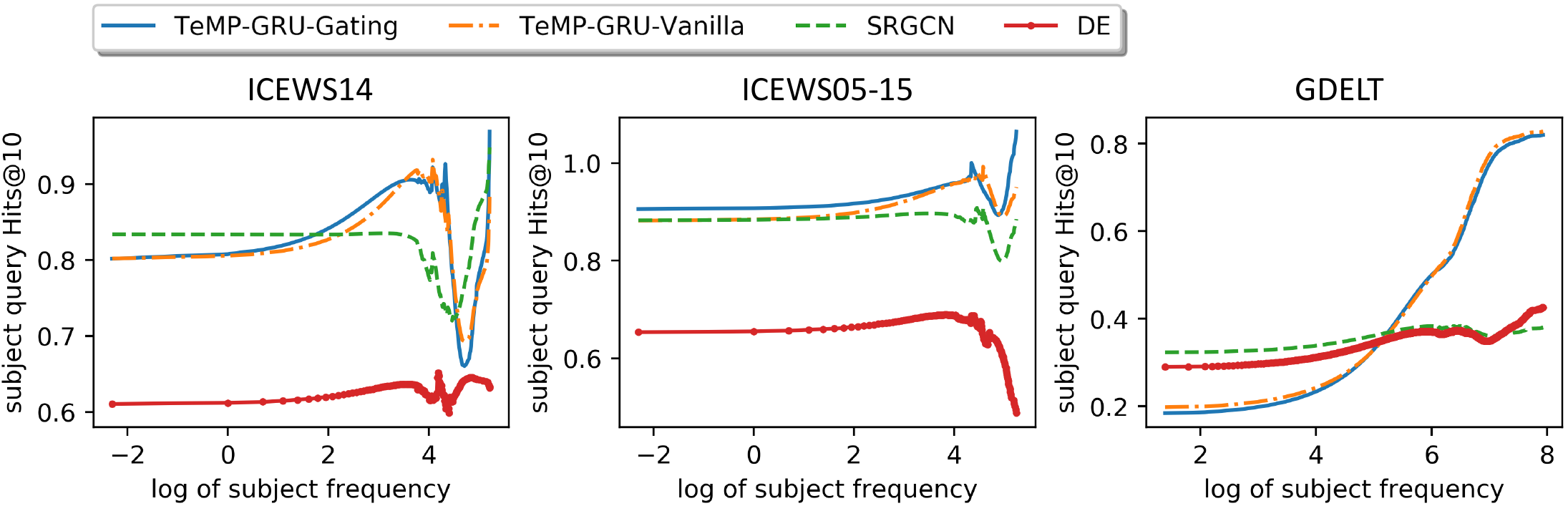}  
\caption{Subject query Hits@10 performances versus temporal subject frequencies}
    \end{subfigure}

    \begin{subfigure}{1\textwidth}
  \includegraphics[width=1\linewidth]{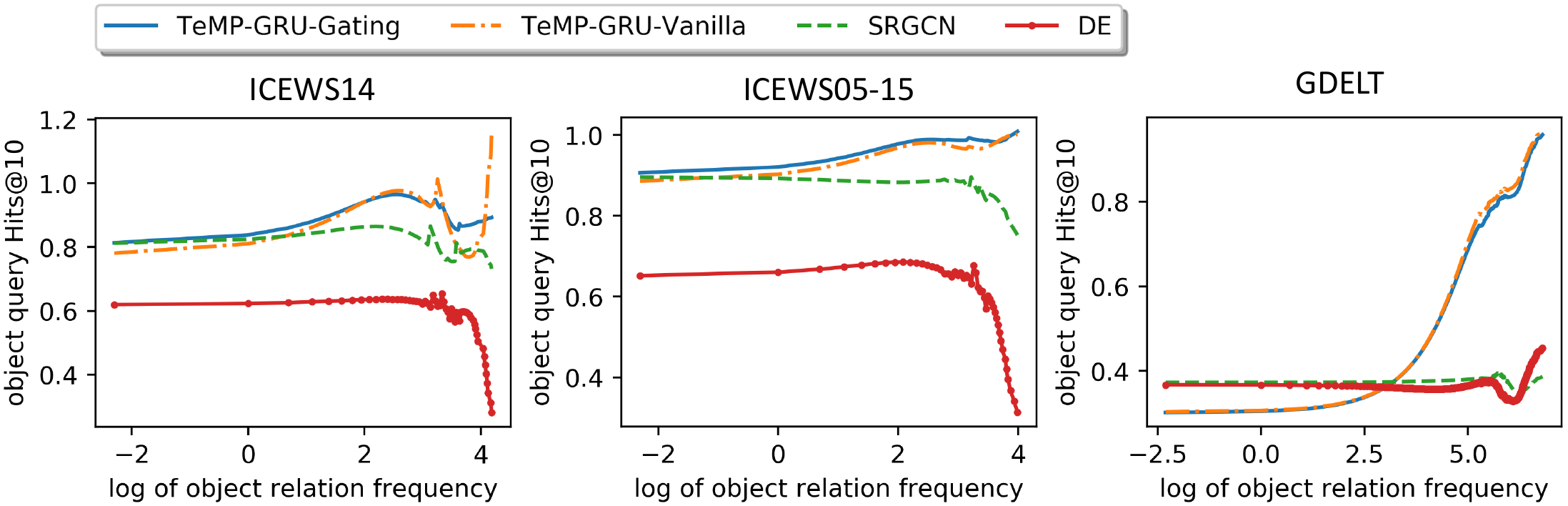}  
\caption{Object query Hits@10 performances versus temporal object-relation frequencies}
    \end{subfigure}
    
    \begin{subfigure}{1\textwidth}
  \includegraphics[width=1\linewidth]{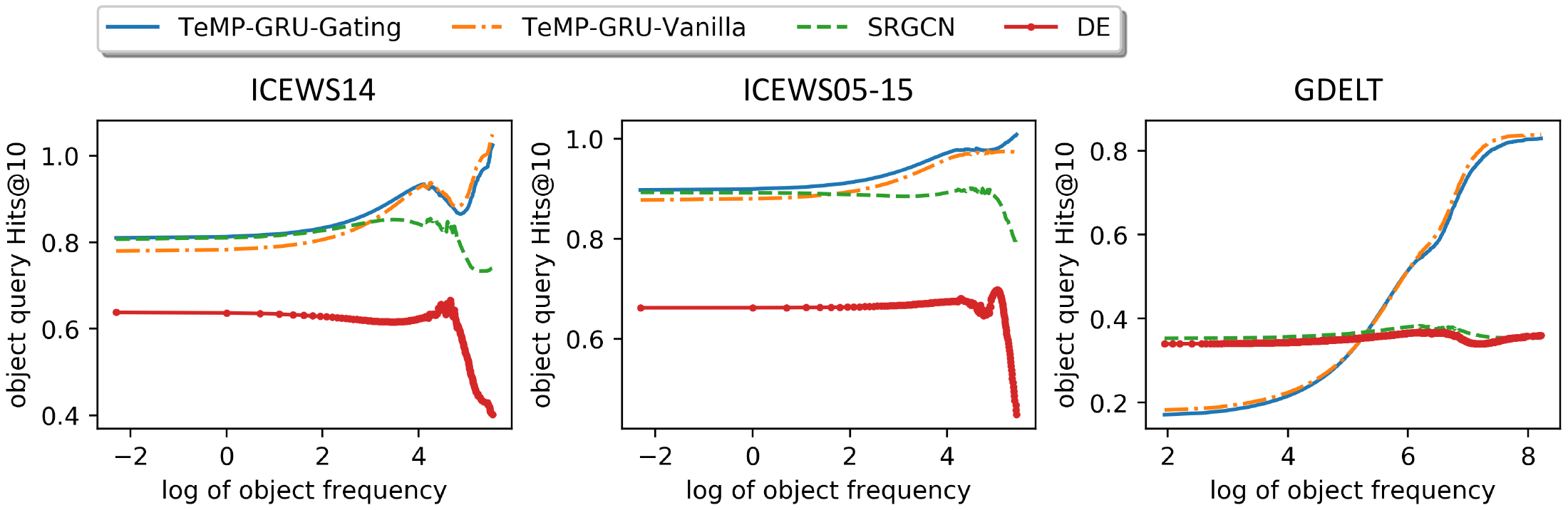}  
\caption{Object query Hits@10 performances versus temporal object frequencies}
    \end{subfigure}
    
    \begin{subfigure}{1\textwidth}
  \includegraphics[width=1\linewidth]{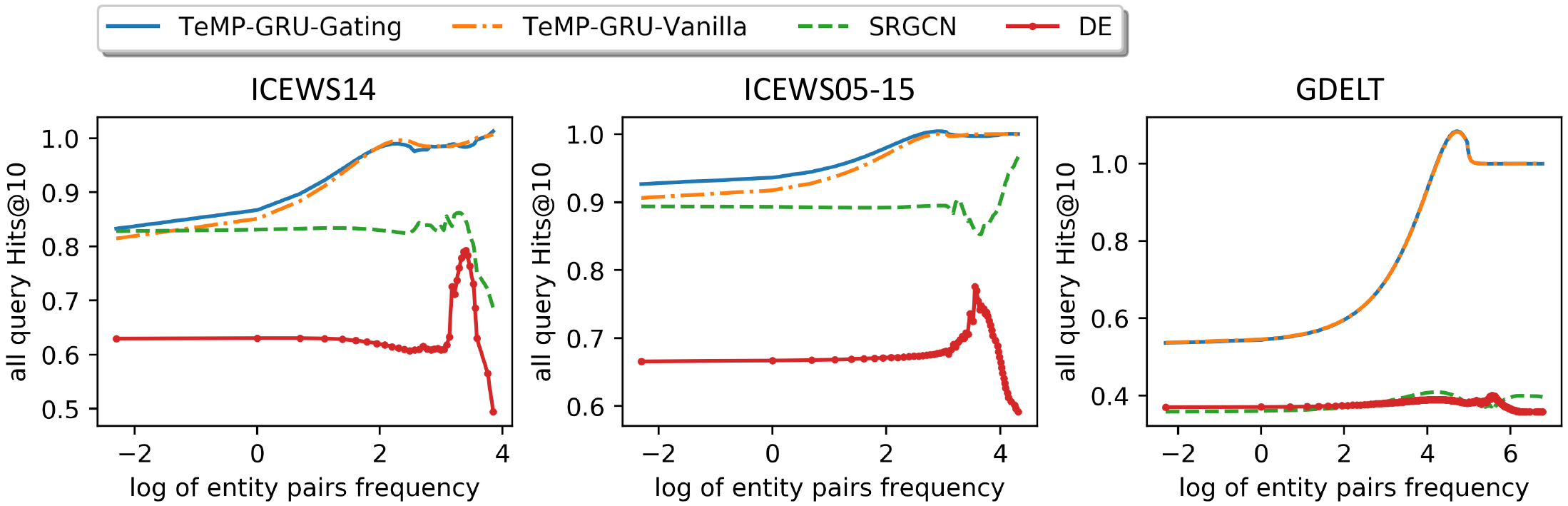}  
\caption{All query Hits@10 performances versus temporal entity pair frequencies}
    \end{subfigure}
\end{figure*}
\begin{figure*}[htb]
    \ContinuedFloat
    \begin{subfigure}{1\textwidth}
  \includegraphics[width=1\linewidth]{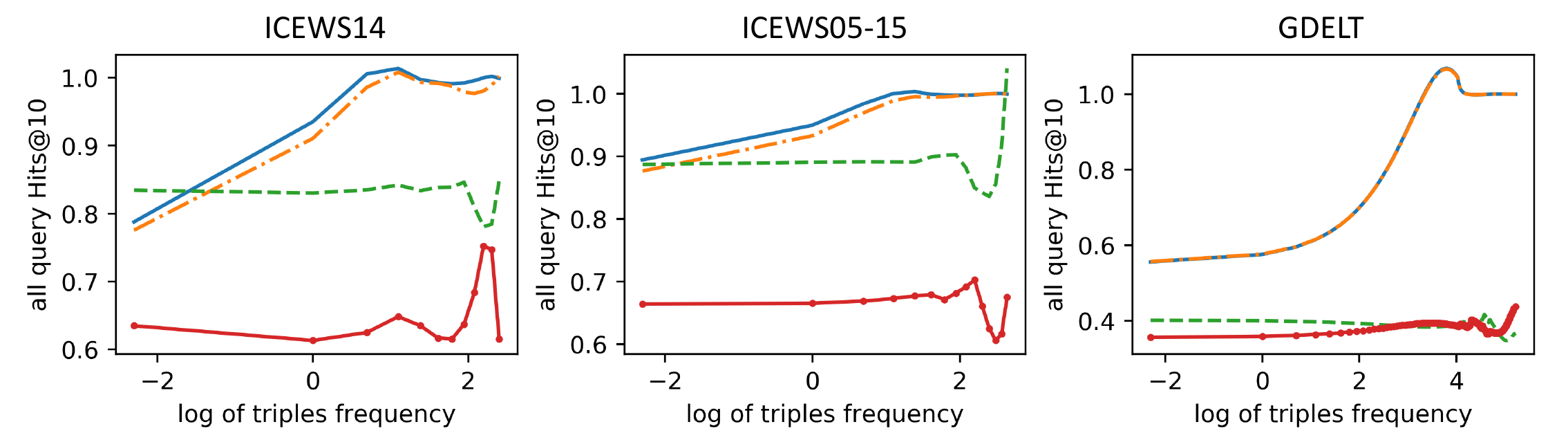}  
\caption{All query Hits@10 performances versus temporal triple frequencies}
    \end{subfigure}
\caption{Plots of replication effect group}
\label{replication figures}
\end{figure*}

% \section{Supplemental Material}
% \subfile{reprodocibility checklist}

\end{document}